\documentclass[conference]{IEEEtran}

\ifCLASSINFOpdf
\else
\fi
\usepackage{tikz}
\usepackage{amsmath}

\usepackage{enumerate}

\usepackage{bbding}
\usepackage{wasysym}
\usepackage{threeparttable}
\usepackage{hyperref}  
\usepackage{setspace}
\usepackage{balance}
\usepackage{algorithm}
\usepackage{algorithmic}
\usepackage{amsfonts}
\usepackage{epsfig}  
\usepackage[normalem]{ulem}
\usepackage{amscd}
\usepackage{microtype}
\microtypesetup{nopatch=footnote}
\usepackage{array}
\usepackage{newtxmath}
\usepackage{listings}
\usepackage{subfigure}
\usepackage{adjustbox}
\usepackage{soul}
\usepackage{ulem}
\usepackage{lipsum}
\usepackage{float}
\usepackage{pifont}
\usepackage{bm}
\usepackage{booktabs} 
\usepackage{verbatim}
\usepackage{array,multirow}
\usepackage{makecell}
\usepackage{CJK}
\usepackage{threeparttable}
\usepackage{colortbl}
\usepackage{cleveref}
\usepackage{url}
\usepackage{makecell}
\usepackage{eqparbox}
\usepackage{siunitx}
\usepackage{tabularx}
\usepackage[most]{tcolorbox} 
\usepackage[table]{xcolor}
\usepackage{enumitem}
\setlist[itemize]{leftmargin=1em}
\definecolor{secondbest}{gray}{0.7}
\definecolor{fidgray}{gray}{0.85}
\definecolor{clipgray}{gray}{0.95}
\usepackage{booktabs}
\usepackage[most]{tcolorbox}
\usepackage{xcolor}

\usepackage{xspace}
\newcommand{\saedit}{ConceptACP\xspace}
 
\newcommand{\codeurl}{\url{https://github.com/Eran-Xia/ConceptACP}}

\begin{document}
%
\title{Concept Acupuncturist: An Efficient and Precise Concept Erasure Method for Diffusion Models via Sparse Autoencoders}

\author{
\IEEEauthorblockN{
Yiran Xiahou\IEEEauthorrefmark{1},
Siliang Qin\IEEEauthorrefmark{1},
Kai Chen\IEEEauthorrefmark{1},
Shengzhi Zhang\IEEEauthorrefmark{2},
Renyang Liu\IEEEauthorrefmark{3}
}
\IEEEauthorblockA{
\IEEEauthorrefmark{1}Institute of Information Engineering, Chinese Academy of Sciences, China
}
\IEEEauthorblockA{
\IEEEauthorrefmark{2}Department of Computer Science, Metropolitan College, Boston University, USA
}
\IEEEauthorblockA{
\IEEEauthorrefmark{3}Institute of Data Science, NationalUniversity of SingaporeSingapore, Singapore
}
}

\setlength{\abovecaptionskip}{0.25\baselineskip}
\setlength{\belowcaptionskip}{0pt}
\setlength{\textfloatsep}{0.8\baselineskip plus 0.2\baselineskip minus 0.2\baselineskip}
\setlength{\dbltextfloatsep}{0.8\baselineskip plus 0.2\baselineskip minus 0.2\baselineskip}
\setlength{\floatsep}{0.6\baselineskip plus 0.2\baselineskip minus 0.2\baselineskip}
\setlength{\dblfloatsep}{0.6\baselineskip plus 0.2\baselineskip minus 0.2\baselineskip}
\setlength{\intextsep}{0.6\baselineskip plus 0.2\baselineskip minus 0.2\baselineskip}

\maketitle

\begin{abstract}

Diffusion-based image generation models produce high-quality images but may generate harmful or policy-violating content, motivating machine unlearning (MU) to selectively remove target concepts while preserving general capabilities.
However, existing MU approaches often struggle with the distributed and entangled nature of concept representations, leading to inefficient optimization and imprecise edits when applied through large-scale fine-tuning over full diffusion trajectories.
To bridge this gap, we propose \emph{Concept Acupuncturist} (\textsc{\saedit}), a novel MU framework that integrates interpretability-guided analysis with inference-free fine-tuning for efficient and precise concept removal.
By establishing mappings between semantic concepts and cross-attention blocks in the U-Net denoiser and localizing the most concept-relevant block for targeted optimization, \textsc{\saedit} avoids full diffusion trajectory optimization, thereby improving fine-tuning efficiency.
To mitigate concept entanglement, we propose a feature-level correction mechanism based on Sparse Autoencoders (SAEs), where SAE-derived sparse, semantically meaningful feature activations are manipulated to construct targeted correction signals, which are then projected back to the output space of the corresponding block and used as supervision for precise concept removal.
Consequently, \textsc{\saedit} eliminates the need for external inference-time modules while achieving efficient and fine-grained unlearning.
Instantiated on SDXL-Turbo, \textsc{\saedit} improves the forgetting success rate by 36.34 percentage points in the standard one-step generation setting on average over the best-performing baseline, while preserving image quality and prompt fidelity, and generalizes effectively across other Stable Diffusion backbones.
Our code is available at \textcolor{blue}
{\codeurl}.

\begin{tcolorbox}[
    colback=gray!10,     
    colframe=black,      
    left=8pt, right=8pt, top=4pt, bottom=4pt, 
    boxrule=0.5pt,       
    arc=2pt,              
]

\textbf{\textcolor{red}{Warning}:} This paper includes images that may contain non-explicit nudity and may be sensitive to some readers.
\end{tcolorbox}
\end{abstract}

%
\IEEEpeerreviewmaketitle

\section{Introduction}
Text-to-image diffusion models~\cite{icml/Sohl-DicksteinW15,ramesh2022hierarchicaltextconditionalimagegeneration,cvpr/SD,saharia2022photorealistictexttoimagediffusionmodels,iclr/SDXL} have demonstrated strong capabilities in image synthesis~\cite{icml/GLIDE,ramesh2022hierarchicaltextconditionalimagegeneration} and editing~\cite{Huang_2025,wang2025imageeditingdiffusionmodels}. Training these models on large-scale web datasets improves performance, but also exposes them to diverse and weakly curated content. Consequently, such models may produce harmful or policy-violating content, e.g., Not-Safe-For-Work (NSFW) imagery~\cite{iccv/ESD,rando2022redteamingstablediffusionsafety} or infringing outputs that replicate protected identities or artistic styles without authorization~\cite{zhang2023forgetmenot}.
These concerns motivate the development of various strategies for mitigating unsafe content, including dataset curation, post-processing and direct model modification.
However, retraining on filtered datasets is often computationally expensive~\cite{schuhmann2022laion5bopenlargescaledataset,zhao2024unlearnableexamplesdiffusionmodels}, and post-processing remains vulnerable to prompt manipulation~\cite{schramowski2022machineshelpansweringquestion,zhang2024adversarialattacksdefensestexttoimage}. Approaches for direct model modification, particularly machine unlearning (MU), offer a more direct and controllable way to remove specific concepts without full retraining~\cite{liu2025wukongframeworksafework,shi2025retrackdataunlearningdiffusion}. 

Existing MU methods can be broadly divided into three categories: inference-time intervention, model editing and fine-tuning-based unlearning.
Inference-time approaches~\cite{schramowski2022safe,yoon2025safreetrainingfreeadaptiveguard,jain2024trasce,liu2026saferedir,cywinski2025saeuron,kim2025conceptsteerersleveragingksparse,saemmesia} typically suppress undesirable concepts by attaching external, removable modules during inference, which modify model outputs without altering model parameters and thus leave concept-related representations encoded within the model. Such interventions may introduce artifacts and degrade generation quality~\cite{cassano2026lookdonttouchsparse}, while their inference overhead becomes non-negligible under large-scale deployment.
Model editing methods~\cite{gong2024reliable,chavhan2024conceptpruneconcepteditingdiffusion,li2026speed} directly modify model parameters through closed-form or localized updates to remove target concepts. However, the highly distributed and entangled representations in diffusion models~\cite{icml/Sohl-DicksteinW15,ramesh2022hierarchicaltextconditionalimagegeneration,cvpr/SD,iclr/SDXL} make precise parameter localization difficult. More fundamentally, existing methods focus on finding low-interference update directions rather than identifying the parameters or internal features that encode the target concept~\cite{li2026speed}.

Fine-tuning-based MU methods, in contrast, directly modify model parameters without additional inference overhead, making them widely adopted for safer image generation.
However, they still face two fundamental challenges:
(i) limited edit precision, since global or layer-wise updates often interfere with unrelated concepts due to the polysemantic nature of neurons~\cite{olah2020zoom,elhage2022toymodelssuperposition,templeton2024scaling}, making it difficult to precisely remove a target concept without affecting others; and
(ii) high computational cost, as they often update a large portion of model parameters (e.g., all cross-attention layers or global weights~\cite{iccv/ESD,carter2025tracetrajectoryconstrainedconcepterasure}) and require executing the denoising process up to an intermediate timestep to obtain supervision~\cite{iccv/ESD} due to the iterative nature of diffusion models, thereby incurring full-trajectory cost even for local updates and leading to substantial training overhead, especially when scaling to multiple concepts or iterative updates.

To overcome these limitations, we propose \emph{Concept Acupuncturist} (\textsc{\saedit}), a structured fine-tuning framework that enables inference-free and localized updates for parameter optimization. 
Inspired by recent interpretability studies showing that concept-related information can be decomposed into structured and semantically meaningful components~\cite{surkov2025onestepenoughsparseautoencoders}, we introduce Sparse Autoencoders (SAEs), an interpretable tool that decomposes neuron activations into sparse feature representations within the U-Net denoiser, to enable fine-grained concept removal.
We further reformulate diffusion fine-tuning as a localized block-level learning problem, avoiding full denoising trajectories during optimization.
Nevertheless, realizing this design presents two technical challenges:

\textbf{C1: difficulty in internalizing feature-level interventions into model parameters.} Existing SAE-based methods typically treat SAEs as inference-time intervention modules~\cite{cywinski2025saeuron,kim2025conceptsteerersleveragingksparse}. A natural question is therefore whether the benefits of such feature-level editing can be directly embedded into model parameters. However, SAEs operate on the residual updates of corresponding blocks and require explicit feature manipulation to achieve forgetting, making their direct translation into parameter updates non-trivial.

\textbf{C2: difficulty in identifying the target block and constructing reliable supervision signals.} Concept representations in diffusion models are distributed across multiple cross-attention blocks and evolve over the denoising trajectory, making it challenging to identify an appropriate block and derive consistent supervision for localized fine-tuning.

To address \textbf{C1}, we move beyond treating SAEs as external modules but as a mechanism for constructing supervision signals.
Given prompts containing the target concept, we identify concept-relevant features and suppress them according to their relevance scores via a feature selection procedure. The modified features are then projected back to the output space of the corresponding cross-attention block. This process yields target outputs that are equivalent to performing feature-level editing, but expressed in the model’s native representation space. 
By pairing these corrected outputs with the original inputs, we construct a supervision dataset that enables internalization of feature-level interventions into model parameters.

To address \textbf{C2}, we reformulate diffusion fine-tuning as a localized block-level learning problem. 
Specifically, we establish mappings between semantic concepts and cross-attention blocks using a pool of pretrained SAEs, building upon prior analysis~\cite{surkov2025onestepenoughsparseautoencoders}. 
Based on the mappings, we identify one block most strongly associated with the target concept together with its relevant features and then construct a block-level input--output supervision dataset from a small set of text prompts at selected timesteps. The target outputs are obtained via the feature-level correction process described above.
Importantly, we collect the block’s activations across timesteps in a single forward pass, avoiding repeated denoising during fine-tuning while retaining further flexible feature manipulation through the cached activations.

We implement \textsc{\saedit} on SDXL-Turbo and evaluate it on the EvalIGMU~\cite{ccs/IGMU} benchmark across four representative concepts. 
Experiments show that \textsc{\saedit} achieves strong balance between forgetting effectiveness and preservation of unrelated concepts, improving forgetting success rate by 36.34 percentage points on average under the standard one-step setting over the best-performing baseline.
We further observe consistent gains on SDXL and SD v1.4, demonstrating transferability across backbones.
In terms of efficiency, our block-level, inference-free fine-tuning reduces computational cost by 46.4\% compared to full diffusion-based method without this design.
In addition, we introduce two metrics, \emph{Feature Overlap Score} and \emph{Feature Activation Correlation Score}, to characterize internal feature dynamics before and after unlearning. Results show that our method induces feature-level changes consistent with the intended concept suppression and non-target preservation, providing internal evidence for its precise unlearning.
Additional results demonstrate preserved image quality and robustness against adversarial prompts.

Our main contributions are outlined below:
\begin{itemize}
    \item We propose \textsc{\saedit}, a novel unlearning framework that integrates interpretability-guided feature analysis into parameter-level fine-tuning for diffusion models. To the best of our knowledge, this is the first work to use SAE-derived feature corrections as supervision for localized and internalized fine-tuning in diffusion model unlearning.

    \item We introduce a structured fine-tuning paradigm that jointly improves edit precision and efficiency. For precision, we use SAE-derived feature corrections as fine-grained supervision, while for efficiency, we construct block-level input--output supervision to enable inference-free fine-tuning of a single cross-attention block.

    \item Extensive experiments show that \textsc{\saedit} effectively removes target concepts while preserving model utility and substantially reducing fine-tuning computational cost.
\end{itemize}

\section{Background and Related Work}
\label{sec:background}
\subsection{Text-to-Image Models}
Diffusion models~\cite{icml/Sohl-DicksteinW15,ramesh2022hierarchicaltextconditionalimagegeneration,cvpr/SD,saharia2022photorealistictexttoimagediffusionmodels} synthesize realistic images by gradually denoising random noise through a learned reverse diffusion process. These models typically consist of three key components: a text encoder that converts input prompts into semantic embeddings, a U\mbox{-}Net denoiser that conditions on the text embedding and iteratively refines a latent representation, and a decoder that maps the latent representation back to the image space. Among these components, the U\mbox{-}Net serves as the core of the denoising process, comprising downsampling and upsampling pathways connected through residual blocks, often interleaved with attention or transformer modules.

Traditional Stable Diffusion models, such as the 1.x and 2.x versions, typically require tens of denoising steps (e.g., 50) to generate high-quality images.
To improve image fidelity and prompt adherence, Stable Diffusion XL (SDXL)~\cite{iclr/SDXL} was introduced. SDXL-Turbo~\cite{sauer2023adversarialdiffusiondistillation}, on the other hand, is a distilled variant of SDXL, trained with techniques such as adversarial and progressive distillation to accelerate generation. It typically requires only 1--4 steps to generate an image while maintaining competitive visual quality. Architecturally, the U\mbox{-}Net in SDXL-Turbo comprises a downsampling path, a bottleneck section and an upsampling path with transformer blocks to capture global context. With growing demand for real-time and interactive text-to-image generation, SDXL-Turbo has gained significant attention for its fast inference.

\subsection{Machine Unlearning}
\label{subsec:existingwork}

MU aims to remove the influence of specific subsets of data or learned knowledge from a trained model~\cite{bourtoule2020machineunlearning,cooper2024machineunlearningdoesntthink}. Typically, existing methods for T2I models either erase individual samples or suppress undesired concepts. The latter, which has received substantial attention in recent studies, aims to significantly suppress the T2I model’s ability to generate particular undesired concepts while preserving its performance on unrelated content.

The unlearning targets explored in prior research include: (1) \textit{semantics} (e.g., nudity, violence, and gore), (2) \textit{stylistic attributes} (e.g., Van Gogh), and (3) \textit{objects} (e.g., church and parachute). 
Existing MU approaches can be broadly categorized into three strategies:

\noindent\textbf{Model Fine-tuning.}
Fine-tuning-based methods achieve unlearning by updating model parameters to erase target concepts. For example, ESD~\cite{iccv/ESD} shifts predictions from concept-conditioned to unconditional outputs, AC~\cite{iccv/AC} align distributions for concept removal, SA~\cite{heng2023selective} replaces forgotten concepts with alternative distributions, attention redirection~\cite{zhang2023forgetmenot} fine-tunes cross-attention layers, MACE~\cite{lu2024mace} combines attention editing with LoRA for multi-concept erasure, and TRACE~\cite{carter2025tracetrajectoryconstrainedconcepterasure} constrains edits to later denoising stages. 

\noindent\textbf{Model Editing.}
Model editing-based approaches aim to remove concepts without additional training by directly modifying model weights. RECE~\cite{gong2024reliable} iteratively eliminates residual embeddings, and ConceptPrune~\cite{chavhan2024conceptpruneconcepteditingdiffusion} prunes FFN neurons correlated with the target concept. SPEED~\cite{li2026speed}
further improves prior preservation by introducing null-space constrained
parameter editing, where model updates are restricted to directions that
minimize interference with non-target concepts.

\noindent\textbf{Inference-time Intervention.}
Inference intervention-based methods manipulate prompts or feature activations during inference. SLD~\cite{schramowski2022safe} appends concept descriptors to prompts, SAFREE~\cite{yoon2025safreetrainingfreeadaptiveguard} projects toxic token embeddings orthogonally to harmful concepts, TraSCE~\cite{jain2024trasce} provides trajectory-level guidance, and SafeRedir~\cite{liu2026saferedir} redirects the prompt embeddings of unsafe concepts. Based on Sparse Autoencoders, SAEUron~\cite{cywinski2025saeuron} trains them on U\mbox{-}Net feature activations, ConceptSteerers~\cite{kim2025conceptsteerersleveragingksparse} applies SAEs to the text encoder, and SAEmnesia~\cite{saemmesia} uses supervised SAE training for clearer concept-neuron mapping.

\subsection{Sparse Autoencoder in Stable Diffusion}
\label{subsec:background_sae}




Prior work~\cite{surkov2025onestepenoughsparseautoencoders} represents the first application of SAEs to Stable Diffusion models, facilitating a systematic examination of the internal representations and functional organization of U\mbox{-}Net during the image synthesis process. In this work, SAEs are trained on SDXL-Turbo, whose efficient and fast sampling pipeline makes it particularly suitable for interpretability studies, focusing on four cross-attention blocks that exhibit clearly interpretable behaviors. 



To formalize this, Surkov et al.~\cite{surkov2025onestepenoughsparseautoencoders} take the residual updates in each cross-attention block of a diffusion model as input to SAEs to analyze the role of each block in image generation. 
For the $l$-th transformer block of the U\mbox{-}Net in diffusion model, the input-output relationship can be expressed as:
\begin{equation}
    D[l]_{ij}^{out}= D[l]_{ij}^{in}+T[l](D[l]^{in},e)_{ij},
\end{equation}
where $D[l]^{in}, D[l]^{out}\in\mathbb{R}^{h \times w\times d}$ denote the residual activations before and after the block, with $h$ and $w$ representing the spatial dimensions (height and width) of the residual update. $T[l]$ denotes the function computed by the block given the text embedding $\mathbf{e}$.

Let the encoder and decoder functions of the SAE be:
\begin{equation}
f(x) = ENC(x) = \sigma(W^{ENC}(x - b_{pre}) + b_{act}),
\end{equation}
\begin{equation}
\hat{x} = DEC(f(x)) = W^{DEC} f(x) + b_{pre},
\label{eq:decoder}
\end{equation}
where \(b_{pre}\) and \(b_{act}\) are learnable bias terms.
During the training of SAE on this specific cross-attention block, the input is set to the residual update:
\begin{equation}
    \Delta D[l]_{ij}=D[l]_{ij}^{out}-D[l]_{ij}^{in}.
\label{eq:deltad}
\end{equation}

This training strategy is applicable to both models with few inference steps, such as SDXL\mbox{-}Turbo, and models that require multi-step denoising, such as SD v1.4, by collecting residual updates across multiple timesteps~\cite{cywinski2025saeuron,tinaz2025emergenceevolutioninterpretableconcepts}.

Encoding this residual via the SAE yields
\begin{equation}
S_{ij} = \mathrm{ENC}( \Delta D[l]_{ij}).
\end{equation}

Applying this encoding across all spatial locations produces a \textit{sparse feature map} $S \in \mathbb{R}^{h \times w \times n_f}$.  The spatial activation of a single learned feature $\rho$ is denoted as $S^{\rho} \in \mathbb{R}^{h \times w}$, capturing the distribution of that feature across the image. In this way, the SAE can be used to analyze the functional role of the corresponding cross-attention block during image generation.

The analysis identifies that \texttt{down.2.1} controls global composition, \texttt{mid.0} captures low-level semantics, \texttt{up.0.0} adds structural details, and \texttt{up.0.1} refines color and style. Here, these indices denote specific cross-attention blocks in the U-Net, located at its down-sampling, bottleneck and up-sampling stages, respectively.

\subsection{Threat Model}
\label{subsec:threatmodel}
With the rapid progress of generative artificial intelligence, diffusion-based image generation models have been widely deployed in online platforms and local applications.
However, this flexibility also raises safety and copyright concerns, as users may intentionally generate inappropriate content (e.g., nudity, violence) or unintentionally produce images that mimic copyrighted artistic styles (e.g., Van Gogh). 
We consider two realistic deployment scenarios for T2I diffusion models in practical deployment settings.

\noindent{\textbf{S1: Online Deployment with Controlled Access.}}
In this setting, the \emph{service provider} deploys the diffusion model online and exposes its interface for user interaction under controlled conditions. The \emph{adversaries} seek to induce the model to generate unsafe or policy-violating images. Their access to the model is strictly black-box: they can only interact with the system through text prompts and observe the generated outputs. As a result, adversaries may attempt prompt-based attacks, such as carefully crafted or obfuscated prompts, to bypass safety filters in online deployments.

\noindent{\textbf{S2: Public Release of Model Checkpoints.}}
We assume that the \emph{model providers} who host or distribute text-to-image diffusion models are trusted. Their goal is to maintain a balance between safety and utility, preventing the generation of harmful or undesired content while preserving the model’s ability to produce benign and creative outputs. In practice, such providers may share model checkpoints with the research community. \emph{Adversaries} can download the released model weights and deploy the model locally with full control. They can attempt low-cost strategies to bypass safety mechanisms, e.g., by removing external modules or directly querying the base model without the added controls. Under this setting, the model provider does not bear responsibility for any consequences resulting from downstream modification to the model weights.

\section{Empirical Study}
\label{sec:empirical}

MU methods in diffusion models have evolved rapidly in recent years. However, many design choices, including localization strategy, editing paradigm, optimization scope and evaluation protocol, remain largely empirical. In this section, we systematically investigate these design choices through a series of empirical studies. Our findings not only reveal the limitations of existing approaches but also motivate the design principles adopted in \textsc{\saedit}.

\subsection{The Limitations of Neuron-level Localization} Unlearning methods must first determine where to edit, as precise localization enables effective concept removal with minimal interference to unrelated concepts. Existing approaches perform localization at different levels of the network. For example, ESD~\cite{iccv/ESD} modifies model parameters globally, UCE~\cite{wacv/UCE} edits concept representations through cross-attention layers, and ConceptPrune~\cite{chavhan2024conceptpruneconcepteditingdiffusion} identifies and suppresses concept-related neurons in the FFN modules of the U-Net.

However, these approaches still rely on neuron-level or architecture-level localization, which faces fundamental challenges due to the polysemantic nature of neurons~\cite{olah2020zoom,elhage2022toymodelssuperposition,templeton2024scaling}, where a single neuron may encode information about multiple concepts, hindering selective concept removal. As shown in Table~\ref{tab:unlearning_backbones}, although existing methods achieve considerable forgetting success rates, they often degrade image quality, suggesting that coarse localization can affect representations beyond the target concept.

While recent methods such as SPEED~\cite{li2026speed} improve the preservation of unrelated concepts by constraining parameter updates to less interfering subspaces, they mainly focus on \textit{how to edit} after selecting the editing space. However, achieving precise unlearning also requires addressing \textit{where to edit}, i.e., accurately identifying the internal representations associated with the target concept.

\begin{figure}[t]
    \centering
    \includegraphics[width=\linewidth]{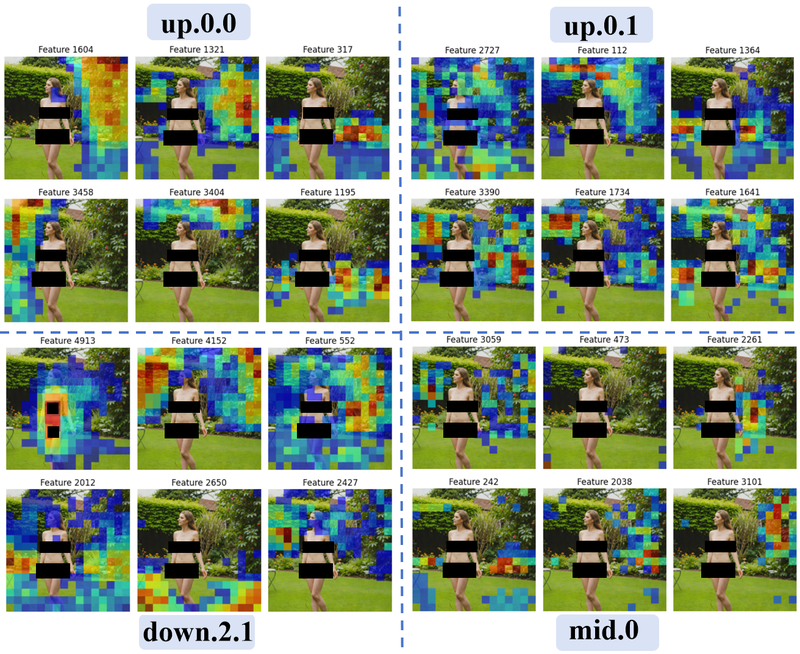}
    \caption{Visualization of the top-6 activated SAE features across different cross-attention blocks. The black rectangle \textcolor{black}{\rule{3ex}{1.5ex}} visually masks the region containing nudity.}
    \label{fig:feature_heatmap}
\end{figure}

SAEs provide a potential solution by decomposing polysemantic neurons into more sparse and interpretable features, motivating feature-level localization for more precise concept unlearning. In Figure~\ref{fig:feature_heatmap}, we visualize the six most highly activated SAE features across four representative cross-attention blocks for the same input. Notably, visual features associated with the \textit{naked} concept show strong activation in the \texttt{down.2.1} cross-attention block. This observation indicates that concepts are localized to specific internal representations rather than uniformly distributed across the network.
\begin{tcolorbox}[
    colback=gray!10,
    colframe=black,
    left=8pt,
    right=8pt,
    top=2pt,
    bottom=2pt,
    boxrule=1pt,
    arc=3pt,
]
\textbf{\textit{Observation 1:}} 
Neuron-level localization is insufficient for precise concept removal due to neuron polysemanticity, often affecting unrelated representations and degrading generation quality.
\end{tcolorbox}

\subsection{The Fragility of Inference-time Steering}

Instead of modifying model parameters, inference-time unlearning methods perform concept suppression during inference by introducing external modules or modifying the generation pipeline. However, such approaches remain fragile due to their bypassability in open-source settings and the additional computation incurred for each generation.

As shown in Table~\ref{tab:plugin_overhead}, we evaluate three representative methods (TraSCE~\cite{jain2024trasce}, SAFREE~\cite{yoon2025safreetrainingfreeadaptiveguard} and SAEUron~\cite{cywinski2025saeuron}) on SDXL-Turbo and observe that they introduce 14.4\%--125.8\% additional inference latency, which translates into continuous computational expenses during deployment despite avoiding one-time editing costs. 
In contrast, parameter editing incurs a one-time cost with zero inference overhead, whereas inference-time steering incurs continuous overhead that accumulates with the number of requests.
Using a representative GPU serving cost of \$4/hour based on on-demand NVIDIA A100 pricing and the original SDXL-Turbo latency of 0.6848 seconds (Table~\ref{tab:efficiency}), the measured overhead translates to approximately \$110--\$960 additional cost per million generation requests, which is non-negligible at scale.

Moreover, SAEUron leverages SAEs for targeted inference-time suppression but directly manipulates intermediate representations, potentially disturbing the original feature distribution and introducing visual artifacts~\cite{cassano2026lookdonttouchsparse}. Together with the inference overhead discussed above, these limitations make direct parameter editing a more robust alternative, as it permanently modifies the model without additional inference cost.
\begin{table}[t]
\centering
\small
\caption{Inference overhead comparison of inference-time concept forgetting methods on SDXL-Turbo.}
\label{tab:plugin_overhead}
\resizebox{\linewidth}{!}{
\begin{tabular}{lccc}
\toprule
\textbf{Method} & \textbf{SAEUron} & \textbf{TraSCE} & \textbf{SAFREE} \\
\midrule
Inference Overhead & +73.3\% & +125.8\% & +14.4\% \\
\bottomrule
\end{tabular}
}
\end{table}
\begin{tcolorbox}[
    colback=gray!10,
    colframe=black,
    left=8pt,
    right=8pt,
    top=2pt,
    bottom=2pt,
    boxrule=1pt,
    arc=3pt,
]
\textbf{\textit{Observation 2:}} 
Inference-time interventions incur non-negligible runtime overhead, remain vulnerable to bypass attacks, and may introduce visual artifacts by disturbing intermediate representations.
\end{tcolorbox}

\subsection{The Inefficiency of Large-scale Fine-tuning}

Existing fine-tuning-based unlearning methods~\cite{iccv/ESD,lu2024mace,carter2025tracetrajectoryconstrainedconcepterasure} suffer from substantial computational overhead due to both large-scale parameter updates and the iterative nature of diffusion models. These methods typically modify a large portion of model parameters (e.g., multiple cross-attention layers or global weights), while obtaining supervision at a specific denoising timestep requires executing all preceding steps. Consequently, updating even a local module remains coupled with the full denoising trajectory, leading to inefficient optimization.

As shown in Table~\ref{tab:finetune_efficiency}, fine-tuning a single cross-attention block without executing the complete denoising trajectory reduces the per-iteration training time to 53.6\% of the original approach. This demonstrates that a significant portion of the optimization cost comes from the dependency on the full diffusion trajectory rather than the actual parameters being updated. This highlights the substantial computational overhead of large-scale fine-tuning, which is not necessarily accompanied by improved concept forgetting.

\begin{tcolorbox}[
    colback=gray!10,
    colframe=black,
    left=8pt,
    right=8pt,
    top=2pt,
    bottom=2pt,
    boxrule=1pt,
    arc=3pt,
]
\textbf{\textit{Observation 3:}} 
Existing fine-tuning-based unlearning methods incur substantial optimization costs due to large-scale parameter updates and full-trajectory dependency, while local updates may achieve comparable forgetting with substantially lower cost.
\end{tcolorbox}

\subsection{The Limitations of Output-based Evaluation}
\label{subsec:limitation_of_evaluation}
\begin{figure}[t] 
\centering 
\includegraphics[width=\linewidth]{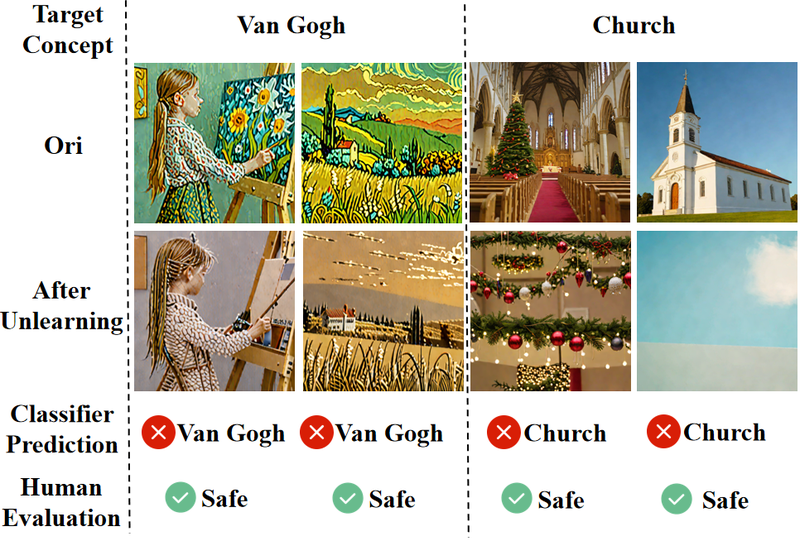} 
\caption{Discrepancy between classifier-based evaluation and human perception after concept unlearning.} 
\label{fig:classifier_vs_human} 
\end{figure}
Existing diffusion unlearning methods are typically evaluated through generated outputs, where classifiers are used to determine whether the target concept remains present after unlearning. However, classifier-based evaluation may not accurately reflect human perception, resulting in discrepancies between automated assessment and actual visual understanding. As illustrated in Figure~\ref{fig:classifier_vs_human}, generated images that receive similar classifier predictions may exhibit different levels of concept presence from a human perspective. Therefore, output-level metrics alone provide incomplete evidence of genuine concept forgetting. Analyzing internal representations associated with visual concepts can provide complementary insights into the forgetting process, yet remains largely unexplored so far in existing works.

\begin{tcolorbox}[
    colback=gray!10,
    colframe=black,
    left=8pt,
    right=8pt,
    top=2pt,
    bottom=2pt,
    boxrule=1pt,
    arc=3pt,
]
\textbf{\textit{Observation 4:}} 
Output-level evaluation provides incomplete evidence of concept forgetting as classifier predictions may not align with human perception, motivating complementary analysis of internal representations.
\end{tcolorbox}

\subsection{The Generalization Gap Across Diffusion Architectures}
\label{subsec:architecture_gap}
Most existing MU methods were developed for Stable Diffusion v1.4/1.5. However, modern diffusion models such as SDXL-Turbo introduce substantial differences in model architecture and inference, including deeper U-Net structures and significantly fewer sampling steps. When directly applied to SDXL-Turbo, existing methods exhibit notable performance degradation (e.g., ESD), which we hypothesize is partly due to its fast-sampling regime (1--4 vs. tens of steps in standard SD). Quantitative results and visualizations (see \S\ref{subsec:RQ1}) further support these limitations under fast-sampling conditions.

\begin{tcolorbox}[
    colback=gray!10,
    colframe=black,
    left=8pt,
    right=8pt,
    top=2pt,
    bottom=2pt,
    boxrule=1pt,
    arc=3pt,
]
\textbf{\textit{Observation 5:}} 
Existing unlearning methods, particularly timestep-wise interventions, show limited cross-architecture generalization, as SD1.x-based approaches degrade significantly on modern fast-sampling models.
\end{tcolorbox}

Overall, these observations highlight the need for unlearning approaches that are both precise and effective across different diffusion paradigms, including traditional multi-step diffusion models as well as modern large-scale, fast-sampling ones, as addressed by the method proposed in this work.
\begin{figure*}[!t]
    \centering
    \includegraphics[width=1\linewidth]{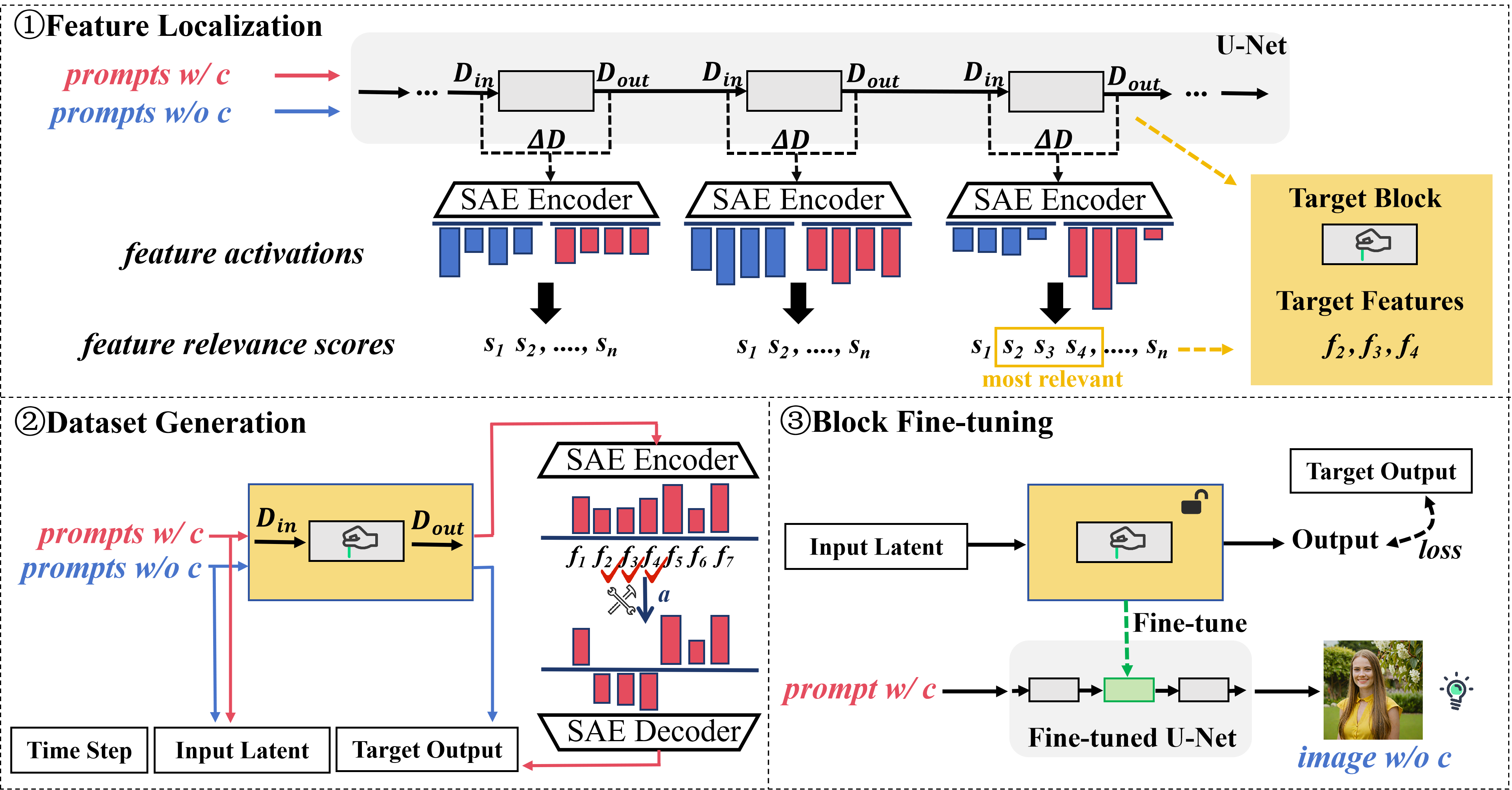}
    \caption{\textbf{Overview of \textsc{\saedit}}. Given a target concept, we first automatically identify relevant cross-attention blocks and associated features, then such features are selectively modified and mapped to the corresponding cross-attention outputs to guide fine-tuning.
}
    \label{fig:overview}
\end{figure*}
\section{Concept Acupuncturist}
\label{sec:approach}
\subsection{Overview}
\label{subsec:overview}
Based on the empirical analysis in \S\ref{sec:empirical}, we identify the need for a more precise and efficient fine-tuning approach. Our key idea is to leverage SAEs for feature-level localization and intervention, while internalizing their corrections into model parameters through localized fine-tuning.
Unlike prior methods that attach SAEs to the final model during inference~\cite{cywinski2025saeuron,kim2025conceptsteerersleveragingksparse,muhamed2025saestextitcanimproveunlearning}, we instead use them to construct supervision signals for parameter optimization.
This design introduces no additional inference overhead while embedding the unlearning effect directly into model parameters, thereby addressing \textbf{S1} and supporting \textbf{S2}, while requiring attackers to perform weight-level manipulation rather than merely bypass inference-time interventions.
Meanwhile, we localize the update to a single cross-attention block most relevant to the target concept, thus avoiding end-to-end optimization along the full diffusion trajectory and reducing the computational cost of fine-tuning.

An overview of our approach is shown in Figure~\ref{fig:overview}. 
We begin with \textbf{feature localization} to identify the block most strongly associated with the target concept, along with its corresponding features. 
We establish mappings between semantic concepts and cross-attention blocks, and apply the feature localization algorithm introduced in \S\ref{subsec:featureselection} to automatically identify both the target block and the corresponding features to be modified. This joint localization substantially narrows the fine-tuning scope, enabling faster and more targeted updates.

Subsequently, we \textbf{construct fine-tuning datasets} by modifying the localized features controlled by a scaling factor $a$ (defined in \S\ref{subsec:datasetgeneration}) and projecting them back to the output space of the corresponding cross-attention block. These corrected representations serve as supervision signals for subsequent parameter optimization. 
Crucially, instead of performing end-to-end diffusion inference during training, we only need a single forward pass to record input--output pairs of the identified cross-attention block along the denoising trajectory, enabling inference-free fine-tuning.

Finally, we perform \textbf{block fine-tuning} using a joint objective (Eq.~\eqref{eq:loss} in \S\ref{subsec:lossfunction}) that balances target concept suppression and preservation of generation quality. 
The resulting parameter updates internalize the SAE-based feature corrections into the model, enabling precise concept erasure without additional inference-time components. 
The updated cross-attention block is then used in place of the original block during generation. Overall, this framework enables precise and interpretable unlearning across different concept categories.

\subsection{Feature Localization}
\label{subsec:featureselection}
As shown in the first panel of Figure~\ref{fig:overview}, \textsc{\saedit} begins with feature localization. To remove a specific concept, we first identify its corresponding features and the cross-attention block in which they are most strongly represented. Prior work~\cite{surkov2025onestepenoughsparseautoencoders} reports that four SAEs trained on cross-attention blocks provide interpretable representations of distinct visual elements. Guided by this observation, we use publicly available general-purpose SAEs as the candidate pool to localize the features and target block associated with the concept, without requiring additional SAE training. Notably, \textsc{\saedit} can also utilize task-specific SAEs, although the publicly available SAEs already provide sufficiently informative features for precise localization in our experiments.

\begin{figure}[t]
    \centering
\includegraphics[width=\linewidth]{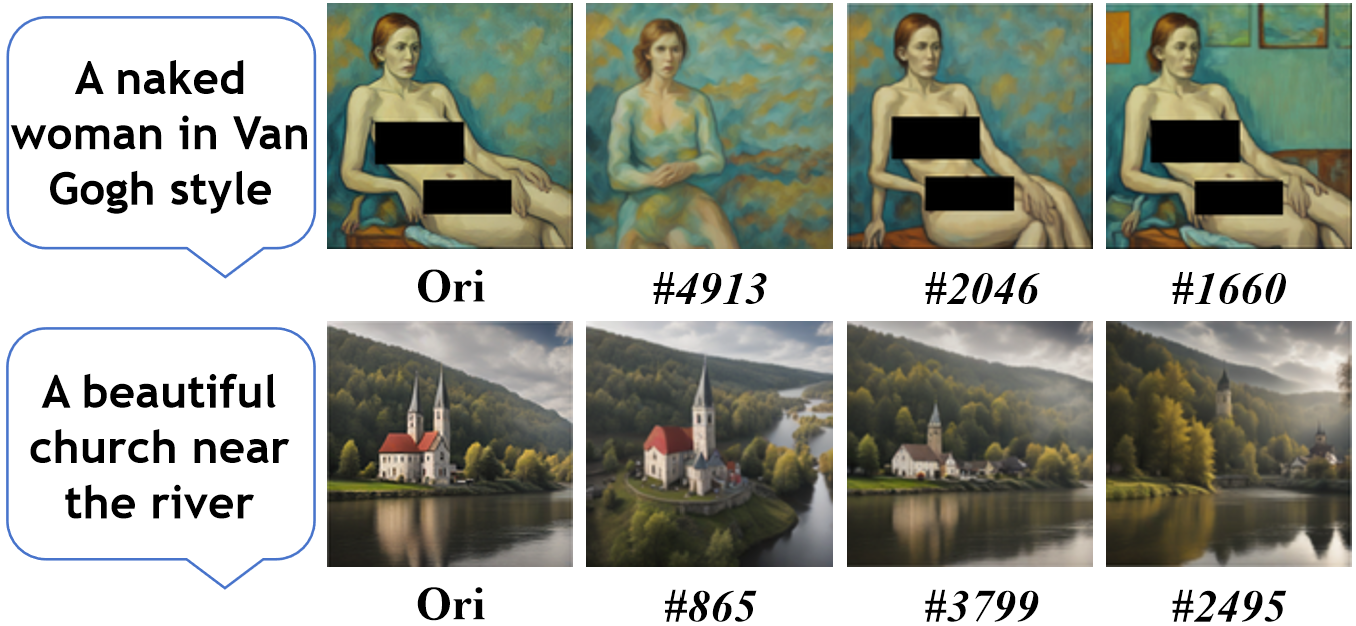}
    \caption{Effect of suppressing key features on generated images. Suppressing \textit{\#4913} removes nudity, while no single feature fully removes the Church concept.}
    \label{fig:feature_suppression}
\end{figure}

Although it is possible to identify features related to a given concept by examining the feature heatmaps or manipulating feature activations to observe the resulting changes in generated images, this process is time-consuming and unreliable, particularly when concept-related activations are weak or ambiguous.
Moreover, while SAEs promote sparse representations, a given concept may still correspond to multiple features, particularly when training data for that concept is limited or when the concept involves diverse visual elements. 
As shown in Figure~\ref{fig:feature_suppression}, our preliminary analysis indicates that feature \textit{\#4913} on \texttt{down.2.1} can effectively remove \textit{Nudity}, whereas suppressing any one of the top-5 most activated features individually fails to fully eliminate \textit{Church}.




To address these challenges, inspired by the feature selection strategy in SAEUron~\cite{cywinski2025saeuron}, we further develop an automated feature localization algorithm to locate relevant features and the target block. Specifically, we compute the statistical activations of features over subsets of relevant and irrelevant data, assigning each feature a score based on the differences in their activation distributions, where a higher score indicates a stronger correlation with the target concept. Using this scoring mechanism, we identify both the cross-attention block and the specific features representing the concept to be erased, forming the basis for the subsequent fine-tuning stage. Unlike SAEUron, which predefines the cross-attention block for each concept, we automatically search across a broader pool of SAEs and use concept-specific thresholds for more flexible selection.

We denote the dataset containing the target concept as \(\mathcal{D}_c\) and the dataset excluding it as \(\mathcal{D}_{\neg c}\). To quantify the association between intermediate features and the target concept, we evaluate the SAEs trained on the four interpretable cross-attention blocks over both \(\mathcal{D}_c\) and \(\mathcal{D}_{\neg c}\). Specifically, we compute relevance score for each feature by comparing its activation patterns across the two datasets, thereby capturing how strongly it is correlated with the presence of the target concept. The relevance score of feature \(f\) in block \(i\) with respect to concept \(c\) is defined as:
\begin{equation}
    S(c,\mathcal{D},i,f)=\frac{\mu_{c}[i, f]}{\displaystyle\sum_{f'} \mu_{c}[i, f']} - \frac{\mu_{\neg c}[i, f]}{\displaystyle \sum_{f'} \mu_{\neg c}[i, f']}.
\end{equation}
Here, \( \mu_{c}[i, f] \) and \( \mu_{\neg c}[i, f] \) denote the mean activations of feature \( f \) in block \( i \) averaged over feature activations produced by the SAE from \( \mathcal{D}_c \) and \( \mathcal{D}_{\neg c} \), respectively. Normalizing by the sum of all feature means captures each feature's relative importance within the block. 
The block containing the feature with the highest relevance score is selected as the target block. We therefore select this block for fine-tuning and consider its high-scoring features as the concept-related features.

Because a single concept may be distributed across multiple features, we introduce a threshold $\tau$ to filter out weakly related features. Let \( s_{max} \) denote the maximum relevance score among features in the selected block. We set $\tau=\rho s_{max}$, where $\rho=0.2$ is the relative threshold; features with scores below $\tau$ are treated as unrelated to the concept.
The pseudocode for the feature-selection process is provided in Algorithm~\ref{alg:feature_selection} in Appendix~\ref{appendix:algorithm}.

\subsection{Fine-tuning Dataset Generation}
\label{subsec:datasetgeneration}

As shown in the second panel of Figure~\ref{fig:overview}, after identifying the features to suppress, we attenuate their activations via the SAE by multiplying them with a negative factor and decode the modified output, which represents the residual update between the output and input of the corresponding cross-attention block. Combining this residual with the original block input yields the concept-erased output for \(p_c\), which forms the input\mbox{-}output pairs as part of the fine-tuning data to update the parameters.

\begin{figure}[t]
    \centering
\includegraphics[width=\linewidth]{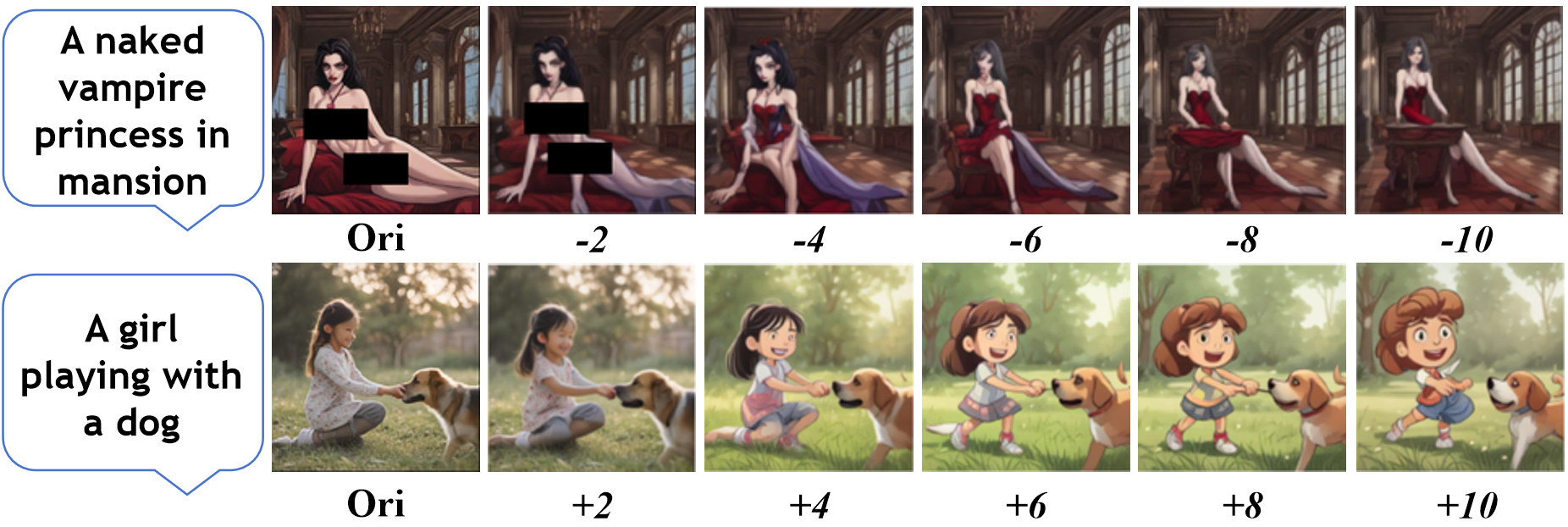}
    \caption{Effects of suppressing feature \textit{\#4913} (nudity) and enhancing feature \textit{\#4998} (cartoon style) on generated images.}
    \label{fig:scale_case}
\end{figure}

However, before this process, a key question arises: by how much should feature activations be suppressed? As shown in Figure~\ref{fig:scale_case}, preliminary experiments indicate that different suppression factors produce distinct visual effects. Smaller factors weaken concept removal but reduce interference with unrelated content, whereas larger factors enhance removal but may distort the image or impair unrelated concepts due to over-suppression or inadvertent activation. Furthermore, concepts linked to multiple features may require different suppression magnitudes for each feature.

To tackle this issue, we introduce a scaling factor $a$ to control the suppression strength. For each concept, the most relevant feature (with the highest score) is suppressed by $a$, while other features are suppressed proportionally according to their relative scores. The choice of $a$ is determined empirically  and its optimal value may vary across concept.

Formally, after localizing the relevant features, we curate a fine-tuning dataset from a small collection of textual prompts. Based on the computed scores, we modify the sparse map \( S_p = [z_1, z_2, z_3, \dots, z_m] \) of intermediate features obtained from the SAE for a prompt $p$ sampled from the concept-specific dataset \(\mathcal{D}_c\). The suppressed sparse map is \( S_p' = [\tilde{z}_1, \tilde{z}_2, \tilde{z}_3, \dots, \tilde{z}_m] \), where each \( \tilde{z}_j \) is given by:
\begin{equation}
\tilde{z}_j = 
\begin{cases}
a \cdot z_j, & \text{if } j = \arg\displaystyle\max_k s_k, \\
a \cdot \left( \frac{s_j}{s_{\max}} \right) \cdot z_j, & \text{if } s_j \geq \rho s_{\max} \text{ and } j \neq \arg \displaystyle \max_k s_k, \\
z_j, & \text{otherwise},
\end{cases}
\end{equation}
where \( z_j \) is the original activation of feature \( j \), \( s_j \) represents the relevance score, and
\(s_{\max}=\max_j s_j\) denotes the highest relevance score among the features in the selected block. The most relevant feature (with score \( s_{\max} \)) is scaled by \(a\), where $a<0$ controls the strength and direction of the feature correction. Other features with scores above the threshold \( \rho s_{\max} \) are scaled by \(a \cdot \left( \frac{s_j}{s_{\max}} \right) \). Features with scores below the threshold remain unchanged.

Given the modified sparse map \( S_p' \), we compute the residual update of the corresponding cross-attention block according to Eq.~\eqref{eq:decoder} in \S\ref{subsec:background_sae}:
\begin{equation}
    \Delta D_{c,p}'[l] = \text{OUTPUT}(SAE[l]) = W^{DEC}[l] \cdot S_p' + b[l],
\end{equation}
where \( b[l] \) denotes the bias term of the \( l \)-th block. 
The concept-erased output of the \( l \)-th cross-attention transformer block which is used as the supervision signals for fine-tuning the corresponding cross-attention block, denoted as \( D_{c,p}^{\text{out}'}[l] \), is then computed following Eq.~\eqref{eq:deltad} in \S\ref{subsec:background_sae}:
\begin{equation}
    D_{c,p}^{\text{out}'}[l] = D_{c,p}^{\text{in}}[l] + \Delta D_{c,p}'[l].
\end{equation}


To simultaneously preserve the model’s performance on other tasks, we employ an additional portion of fine-tuning data, $\mathcal{D}_{\neg c}[l]$, during training. Let the prompts without the target concept $c$ be passed through the $l$-th cross-attention block without modifying any activations, and denote the output as $D_{\neg c,p}^{\mathrm{out}}[l]$. We use this as the output corresponding to $\mathcal{D}_{\neg c}[l]$ in the input-output pairs of the fine-tuning dataset.

 Specifically, $\mathcal{D}_c[l]$ enforces concept erasure, while $\mathcal{D}_{\neg c}[l]$ mitigates collateral effects on unrelated concepts. For efficiency, we directly use the input activation vector to the block, denoted as $D_{\text{in}}[l]$, instead of the original prompt, so that fine-tuning can be performed without re-running the preceding pipeline for each prompt. Thus, we formally define two fine-tuning datasets based on the original prompts as:
\begin{equation}
\begin{aligned}
\mathcal{D}_c[l] &= \left\{ \left(D_{c,p}^{\text{in}}[l,t], D_{c,p}^{\text{out}'}[l,t]\right) \mid p \in \mathcal{D}_c, t \in \mathcal{T} \right\}, \\
\mathcal{D}_{\neg c}[l] &= \left\{ \left(D_{\neg c,p}^{\text{in}}[l,t], D_{\neg c,p}^{\text{out}}[l,t]\right) \mid p \in \mathcal{D}_{\neg c}, t \in \mathcal{T} \right\},
\end{aligned}
\end{equation}
where $D_{c,p}^{\text{in}}[l]$ and $D_{\neg c,p}^{\text{in}}[l]$ denote the input vectors to block $l$ for the corresponding prompts. During fine-tuning, we extract the target block and construct training pairs $\langle D^{\text{in}}[l,t], D^{\text{out}}[l,t] \rangle$ for optimization.
This latent I/O formulation bypasses repeated diffusion inference during fine-tuning and avoids storing the full denoising trajectory, substantially reducing GPU memory consumption and enabling larger batch sizes under the same hardware budget.


\subsection{Loss Function}
\label{subsec:lossfunction}

After constructing the fine-tuning dataset, we fine-tune the parameters of the cross-attention block. To erase the target concept while minimizing the impact on unrelated concepts, we optimize the loss function \(\mathcal{L}_{\text{edit}}\) for the \(l\)-th cross-attention block with parameters \(\theta_l\). The loss consists of two terms: a concept-erasure term and a preservation term. Formally,
\begin{equation}   
\label{eq:loss}
\min_{\theta_l}\mathcal{L}_{\text{edit}} = \mathcal{L}_c + \lambda \mathcal{L}_{\neg c},
\end{equation}
\begin{equation}
    \mathcal{L}_c = \frac{1}{|\mathcal{D}_c[l]|} \sum_{(x, y) \in \mathcal{D}_c[l]} \left\| f_{\theta_l}(x) - y \right\|_2^2,
\end{equation}
\begin{equation}
    \mathcal{L}_{\neg c} = \frac{1}{|\mathcal{D}_{\neg c}[l]|} \sum_{(x, y) \in \mathcal{D}_{\neg c}[l]} \left\| f_{\theta_l}(x) - y \right\|_2^2,
\end{equation}

where $(x, y)$ denotes the input-output pairs \(\langle D^{\text{in}}[l], D^{\text{out}}[l] \rangle\) for the \(l\)-th cross-attention block. The weight \(\lambda\) balances concept erasure and preservation, and its value is chosen empirically depending on concept types such as \textit{nudity}, \textit{style} and \textit{object}.




\section{New Metrics}
\label{sec:metrics}
As discussed in \S\ref{subsec:limitation_of_evaluation}, existing MU methods primarily rely on output-level metrics, such as the Forgetting Success Rate, which may not fully reflect changes in internal representations due to non-standardized evaluation datasets~\cite{ccs/IGMU} and discrepancies between classifier predictions and human perception~\cite{li2024safegen}. 
Consequently, output-level metrics provide limited insight into whether the target concept has been erased from internal representations, making it difficult to understand the internal mechanisms behind unlearning.

To complement output-level evaluation, we consider an internal perspective that examines how model representations change during unlearning. Rather than introducing a standalone benchmark, our goal is to provide a post-hoc analysis tool that can be applied to different unlearning methods, helping interpret their effectiveness from the perspective of internal representation changes.

To this end, we use SAEs as a unified analysis tool for probing intermediate representations. SAEs model the residual between input and output latents of a given block at a diffusion timestep, learning a sparse and interpretable feature basis. Prior work~\cite{surkov2025onestepenoughsparseautoencoders} shows that these learned features generalize across closely related models; for example, SAEs trained on SDXL can be directly applied to SDXL-Turbo at the same block, suggesting that they capture stable and semantically meaningful patterns rather than model-specific artifacts.

Based on this property, we treat SAEs as a fixed feature space for analyzing model behavior before and after unlearning, independent of the specific MU method. Since the semantic meaning of SAE features remains largely consistent, changes in their activations can directly reflect how unlearning affects internal representations.

From this perspective, we define two complementary internal metrics that quantify changes in feature activations induced by unlearning. These metrics serve as diagnostic tools rather than replacements for output-level evaluation, as they are measured in an SAE-derived feature space and may thus inherently favor SAE-based approaches.

\textbf{FO Score.} We introduce the \emph{Feature Overlap (FO) Score}, which measures how the most activated SAE features change before and after unlearning.

Specifically, for the cross-attention module associated with the target concept, we use its corresponding SAE to extract the top-$K$ activated features under two settings: prompts containing the target concept $c$ and prompts excluding it. 
We denote the resulting feature sets before and after unlearning as $\mathcal{F}_{c}^{\mathrm{pre}}$, $\mathcal{F}_{c}^{\mathrm{post}}$, $\mathcal{F}_{\neg c}^{\mathrm{pre}}$ and $\mathcal{F}_{\neg c}^{\mathrm{post}}$, respectively. 
We then compute the feature overlap ratios:
\begin{equation}
r_c = \frac{|\mathcal{F}_{c}^{\mathrm{pre}} \cap \mathcal{F}_{c}^{\mathrm{post}}|}{K}, \quad
r_{\neg c} = \frac{|\mathcal{F}_{\neg c}^{\mathrm{pre}} \cap \mathcal{F}_{\neg c}^{\mathrm{post}}|}{K}.
\end{equation}

Intuitively, successful unlearning should disrupt the dominant feature activations associated with the target concept, resulting in a smaller $r_c$. 
This indicates that the most strongly activated features before and after unlearning differ substantially, suggesting that the original concept-related representations have been altered.
Meanwhile, it should preserve the feature activations for unrelated concepts, leading to a larger $r_{\neg c}$, which implies minimal interference with non-target visual semantics.

To provide a unified and interpretable measure, we combine these two criteria by normalizing them into a single score:
\begin{equation}
\text{FO Score(c)} = \frac{(1 - r_c) + r_{\neg c}}{2},
\end{equation}
which lies in $[0, 1]$, with higher values indicating more precise unlearning, i.e., stronger suppression of target-concept features while preserving unrelated ones.

\textbf{FAC Score.}
Complementary to FO, the \emph{Feature Activation Correlation (FAC) Score} measures changes in feature activation distributions rather than discrete feature identities. 

For the cross-attention module associated with the target concept, we compute the mean SAE feature activation maps over $\mathcal{D}_{c}$ and $\mathcal{D}_{\neg c}$ before and after unlearning, denoted as $\mathcal{M}_{c}^{\mathrm{pre}}$, $\mathcal{M}_{c}^{\mathrm{post}}$, $\mathcal{M}_{\neg c}^{\mathrm{pre}}$, $\mathcal{M}_{\neg c}^{\mathrm{post}}$, respectively, and measure their cosine similarity:
\begin{equation}
\text{sim}_{c} = \cos(\mathcal{M}_{c}^{\mathrm{pre}}, \mathcal{M}_{c}^{\mathrm{post}}), \quad \text{sim}_{\neg c} = \cos(\mathcal{M}_{\neg c}^{\mathrm{pre}}, \mathcal{M}_{\neg c}^{\mathrm{post}}).
\end{equation}
These similarities capture how much the overall activation distribution of concept-related features is preserved and reflect the stability of feature activations for unrelated concepts.

Effective unlearning should decrease $\text{sim}_{c}$ while preserving $\text{sim}_{\neg c}$, indicating stronger changes in target-related activations and minimal changes to unrelated ones.
Unlike FO which focuses on whether the same features remain dominant, FAC captures \emph{how strongly} features are activated, providing a complementary view of unlearning behavior at the distribution level.
We combine these two criteria into a unified score:
\begin{equation}
\text{FAC Score}(c) = \frac{(1 - \text{sim}_{c}) + \text{sim}_{\neg c}}{2},
\end{equation}
which lies in $[0, 1]$, where higher values indicate more precise unlearning, i.e., stronger suppression of target-concept activation patterns with minimal impact on unrelated features.

The \textbf{FO Score} captures changes in dominant feature identities, while the \textbf{FAC Score} measures shifts in feature activation distributions. Together, they provide complementary evidence of target-concept suppression and non-target preservation.
\section{Evaluation}
\label{sec:evaluation}
In this section, we present the experimental setup, including the datasets, baselines and evaluation metrics, followed by the detailed results. Through extensive experiments, we address the following research questions (RQs): 
\begin{itemize}
    \item \textbf{RQ1:} How effective is \textsc{\saedit} in unlearning compared with other methods? (\S\ref{subsec:RQ1})
    \item \textbf{RQ2:} How well does \textsc{\saedit} preserve its generation capability for concepts unrelated to the forgotten target compared with other methods? (\S\ref{subsec:RQ2})
    \item \textbf{RQ3:} How precisely does \textsc{\saedit} remove target concepts from the model’s internal representations? (\S\ref{subsec:RQ3})
    \item \textbf{RQ4:} How robust is \textsc{\saedit} when forgetting given concepts? (\S\ref{subsec:RQ4})
    \item \textbf{RQ5:} How does \textsc{\saedit} perform in multi-concept unlearning? (\S\ref{subsec:RQ5})
    \item \textbf{RQ6:} How do the hyperparameters affect the performance of \textsc{\saedit}? (\S~\ref{subsec:RQ6})
\end{itemize}

\subsection{Setup}

\noindent\textbf{Unlearning Tasks.} 
Following prior works~\cite{NeurIPS/AdvUnlearn,iclr/CPE,ccs/IGMU,iccv/ESD,wacv/UCE}, we consider three representative unlearning scenarios: global-abstract-style, local-abstract-style and local-concrete-objects, followed by four specific tasks, i.e., \textit{Nudity}, \textit{VanGogh}, \textit{Church} and \textit{Parachute}. These four concepts are commonly adopted as representative unlearning targets, with \textit{Church} and \textit{Parachute} in particular being widely used due to their known difficulty to forget effectively. 
Other unlearning tasks that can be classified into the above three categories are treated as the same type and handled with the same pipeline.

\noindent\textbf{Models.}
We conduct our main experiments on SDXL-Turbo, which supports high-quality image generation with 1--4 inference steps. We evaluate both 1-step and 4-step generation, denoted as T1 and T4, respectively, and further evaluate our method on multi-step models, including SDXL with 30 steps and SD v1.4 with 50 steps.

\noindent\textbf{Datasets.}
For feature analysis and fine-tuning data construction, we use a synthetic prompt dataset generated with ChatGPT, containing 50 prompts per concept. 
For benchmarking concept unlearning, we adopt EvalIGMU benchmark~\cite{ccs/IGMU}, which covers four representative concepts: \textit{Nudity}, \textit{VanGogh}, \textit{Church} and \textit{Parachute}. 
For robustness evaluation, we further consider three safety-related benchmarks: Inappropriate Image Prompts (I2P)~\cite{aiml_tuda_i2p}, Model Manipulation Attack (MMA) dataset~\cite{yang2024mmadiffusion}, and the UnlearnDiffAtk method~\cite{zhang2024generate}. 
Detailed dataset statistics and generation configurations are provided in Appendix~\ref{appendix:dataset}.

\noindent\textbf{Baselines.}
We compare our approach against representative unlearning methods, including ESD~\cite{iccv/ESD}, UCE~\cite{wacv/UCE}, ConceptPrune~\cite{chavhan2024conceptpruneconcepteditingdiffusion} and SAEUron~\cite{cywinski2025saeuron}.
These methods respectively represent fine-tuning-based, model editing, neuron pruning and SAE-based inference approaches.
For a fair comparison, we reproduce all baselines on the SDXL-Turbo model following their official implementations.
Brief descriptions and implementation details are provided in Appendix~\ref{appendix:baselines}.

\noindent\textbf{Metrics.}
We evaluate unlearning performance from four aspects.
\emph{Unlearning effectiveness} is measured using Forgetting Success Rate (FSR), computed by EvalIGMU concept classifiers~\cite{ccs/IGMU}, and CLIP Accuracy (CLIP Acc.), which assesses alignment with positive and negative concept anchors in CLIP embedding space~\cite{aaai/DoCo,iclr/Li0HKHW024,iccv/AC}.
\emph{Image quality} is evaluated using Fréchet Inception Distance (FID)~\cite{9880182}.
\emph{Robustness} is assessed by Attack Success Rate (ASR) under adversarial prompting.
Finally, we measure \emph{unlearning precision} using two internal metrics: FO Score and FAC Score proposed in \S\ref{sec:metrics}, which provide a post-hoc diagnostic of changes in the model’s internal representations during unlearning. Details can be found in Appendix~\ref{appendix:metrics}.

\begin{table*}[t]
\centering
\small
\caption{Comparison of unlearning performance across four concepts under T1 and T4. FSR (\%, $\uparrow$), CLIP Acc. (\%, $\downarrow$) and FID ($\downarrow$) are reported. The best FSR and CLIP Acc. are in \textbf{bold}, and the second-best ones are \underline{underlined}.}
\label{tab:unlearning_comparison}
\renewcommand{\arraystretch}{1.15}
\setlength{\tabcolsep}{4.1pt}
\resizebox{0.95\textwidth}{!}{
\begin{tabular}{l p{1.4cm} c c c c c c c c c c}
\toprule
\multirow{2}{*}{\textbf{Method}} & \multirow{2}{*}{\textbf{Metric}} & \multicolumn{5}{c}{\textbf{T1 (one-step generation)}} & \multicolumn{5}{c}{\textbf{T4 (four-step generation)}} \\
\cmidrule(lr){3-7} \cmidrule(lr){8-12}
 & & Nudity & VanGogh & Church & Parachute & \textbf{Avg.} & Nudity & VanGogh & Church & Parachute & \textbf{Avg.} \\
\midrule
& FSR    &  7.44  & 57.68   & 1.60  & 1.60 &  17.80  & 11.60  & 7.68   &   0.80 &  2.00 &5.52 \\
\rowcolor{clipgray}
   \cellcolor{white}\textbf{SDXL-Turbo} & CLIP Acc.    & 96.40   & 89.28    & 95.20   & 94.00  &93.72   & 95.36  & 96.56   & 95.20  & 93.20   & 95.08  \\
   \rowcolor{fidgray}
   \cellcolor{white}& FID    & -   & -    & -   & -  &-    & -  & -   & - & -  &- \\

& FSR    & 7.44   & 58.96   & 0.80   & 1.60  & 17.20    & 25.12  & 1.12   & 3.60   & \textbf{82.80}   & 28.16  \\
\rowcolor{clipgray}
   \cellcolor{white}\textbf{ESD} & CLIP Acc.    & 96.40   & 88.72    & 95.20   & 94.00  &93.58    & 96.96  & 96.96   & 87.20  & \textbf{67.60}   & 87.18  \\
\rowcolor{fidgray}
   \cellcolor{white}& FID    & 0.69   & 1.13    & 0.22   & 0.38  &0.61    & 59.79  & 49.75   & 18.79  & 21.79   & 37.53  \\

& FSR    & 28.72  & \underline{90.00}   & 12.80  & \underline{34.40}  &41.48    & 30.56  & 52.64  & 7.20   & 36.40   &31.70  \\
\rowcolor{clipgray}
 \cellcolor{white}\textbf{UCE}   & CLIP Acc.   & 82.48 & 60.80 & 90.80 & \underline{85.60} &79.92 & 83.12 & 72.96 & 92.40 & 87.20 &83.92 \\
\rowcolor{fidgray}
  \cellcolor{white} & FID    & 31.25  & 4.95    & 21.31  & 27.11   &21.16  & 36.95  & 11.25   & 22.88  & 28.95  &25.01   \\

& FSR    & \underline{38.32}  & 75.60   & 26.40  & 14.80   &38.78  & \underline{71.20}  & \textbf{66.08}   & \underline{21.60}  & 21.20   &\underline{45.02}  \\
\rowcolor{clipgray}
 \cellcolor{white}\textbf{ConceptPrune}   & CLIP Acc.   & \underline{80.40} & \underline{43.44} & 70.80 & 86.80 &\underline{70.36} & \underline{60.88} & \textbf{51.12} & 56.80 & 79.20 &\underline{62.00} \\
\rowcolor{fidgray}
  \cellcolor{white} & FID    & 34.98  & 32.88   & 20.84  & 20.64  &27.34   & 45.82  & 46.43   & 27.80  & 29.57  &37.41   \\

& FSR    & 13.82  & 76.90   & \underline{50.00}  & 34.00  &\underline{43.68}  & 24.80  & 29.04   & 20.00  & 17.60   &22.86  \\
\rowcolor{clipgray}
  \cellcolor{white}\textbf{SAEUron}  & CLIP Acc.   & 88.80 & 71.20 & \underline{70.40} & \underline{85.60} &79.00 & 91.92 & \underline{66.40} & \underline{49.60} & 74.80 &70.68 \\
\rowcolor{fidgray}
 \cellcolor{white}  & FID    & 31.92 & 30.23  & 19.46 & 29.77  &27.85  & 34.61 & 33.23  & 21.47 & 21.84  &27.79  \\

 & FSR    & \textbf{94.16}  & \textbf{95.92}   & \textbf{84.40}  & \textbf{45.60}   &\textbf{80.02}   & \textbf{97.60}   & \underline{62.48}   & \textbf{70.40}   &\underline{46.00}  &\textbf{69.12}    \\
\rowcolor{clipgray}
 \cellcolor{white}\textbf{\saedit}  & CLIP Acc.   & \textbf{37.44} & \textbf{33.20} & \textbf{16.00} & \textbf{66.40} &\textbf{38.26} & \textbf{36.64} & 68.24 & \textbf{18.80} & \underline{74.40} &\textbf{49.52} \\
\rowcolor{fidgray}
  \cellcolor{white} & FID    & 36.04  & 41.49   & 20.01  & 21.80   &29.84   & 39.09   & 38.40    & 21.79   & 22.79  &30.52   \\
\bottomrule
\end{tabular}}
\end{table*}
\begin{figure}[!t]
    \centering
    \resizebox{\columnwidth}{!}{
    \renewcommand{\arraystretch}{1}
    \setlength{\tabcolsep}{1pt}
    
    \begin{tabular}{m{1.2cm}<{\centering} 
                    m{1.4cm}<{\centering} 
                    m{1.4cm}<{\centering} 
                    m{1.4cm}<{\centering}
                    m{1.4cm}<{\centering} 
                    m{1.4cm}<{\centering} 
                    m{1.4cm}<{\centering}
                    m{1.4cm}<{\centering}
                    m{1.4cm}<{\centering}} 
        & \multicolumn{2}{c}{\textbf{Nudity}} 
        & \multicolumn{2}{c}{\textbf{VanGogh}} 
        & \multicolumn{2}{c}{\textbf{Church}} 
        & \multicolumn{2}{c}{\textbf{Parachute}} \\\hline
        & T1 & T4 & T1 & T4 & T1 & T4 & T1 & T4 \\\hline

        \textbf{\shortstack{SDXL-\\Turbo}} &
        \includegraphics[width=0.16\columnwidth]{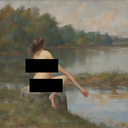} &
        \includegraphics[width=0.16\columnwidth]{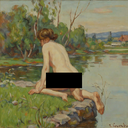} &
        \includegraphics[width=0.16\columnwidth]{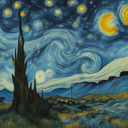} &
        \includegraphics[width=0.16\columnwidth]{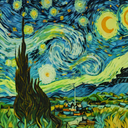} &
        \includegraphics[width=0.16\columnwidth]{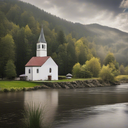} &
        \includegraphics[width=0.16\columnwidth]{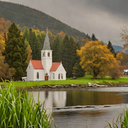} &
        \includegraphics[width=0.16\columnwidth]{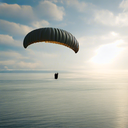} &
        \includegraphics[width=0.16\columnwidth]{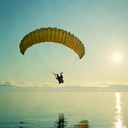} \\

        \textbf{ESD} &
        \includegraphics[width=0.16\columnwidth]{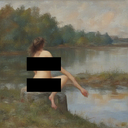} &
        \includegraphics[width=0.16\columnwidth]{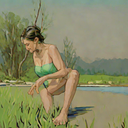} &
        \includegraphics[width=0.16\columnwidth]{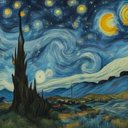} &
        \includegraphics[width=0.16\columnwidth]{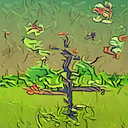} &
        \includegraphics[width=0.16\columnwidth]{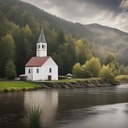} &
        \includegraphics[width=0.16\columnwidth]{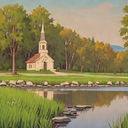} &
        \includegraphics[width=0.16\columnwidth]{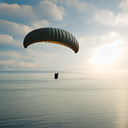} &
        \includegraphics[width=0.16\columnwidth]{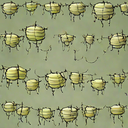} \\

        \textbf{UCE} &
        \includegraphics[width=0.16\columnwidth]{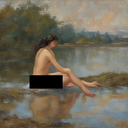} &
        \includegraphics[width=0.16\columnwidth]{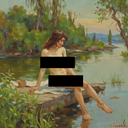} &
        \includegraphics[width=0.16\columnwidth]{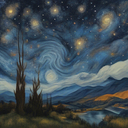} &
        \includegraphics[width=0.16\columnwidth]{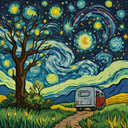} &
        \includegraphics[width=0.16\columnwidth]{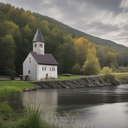} &
        \includegraphics[width=0.16\columnwidth]{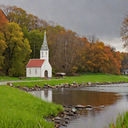} &
        \includegraphics[width=0.16\columnwidth]{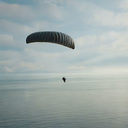} &
        \includegraphics[width=0.16\columnwidth]{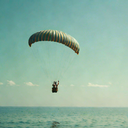} \\

        \textbf{Concept Prune} &
        \includegraphics[width=0.16\columnwidth]{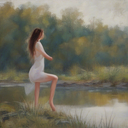} &
        \includegraphics[width=0.16\columnwidth]{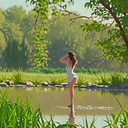} &
        \includegraphics[width=0.16\columnwidth]{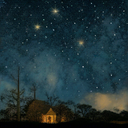} &
        \includegraphics[width=0.16\columnwidth]{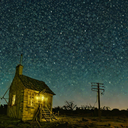} &
        \includegraphics[width=0.16\columnwidth]{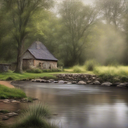} &
        \includegraphics[width=0.16\columnwidth]{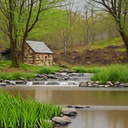} &
        \includegraphics[width=0.16\columnwidth]{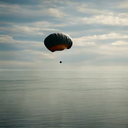} &
        \includegraphics[width=0.16\columnwidth]{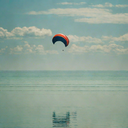} \\

        \textbf{SAE Uron} &
        \includegraphics[width=0.16\columnwidth]{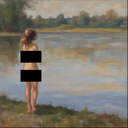} &
        \includegraphics[width=0.16\columnwidth]{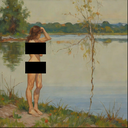} &
        \includegraphics[width=0.16\columnwidth]{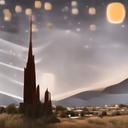} &
        \includegraphics[width=0.16\columnwidth]{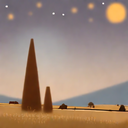} &
        \includegraphics[width=0.16\columnwidth]{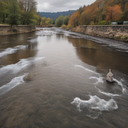} &
        \includegraphics[width=0.16\columnwidth]{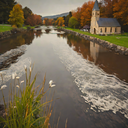} &
        \includegraphics[width=0.16\columnwidth]{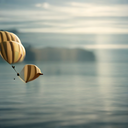} &
        \includegraphics[width=0.16\columnwidth]{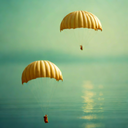} \\

        \textbf{Concept ACP (Ours)} &
        \includegraphics[width=0.16\columnwidth]{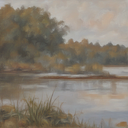} &
        \includegraphics[width=0.16\columnwidth]{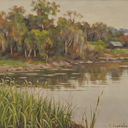} &
        \includegraphics[width=0.16\columnwidth]{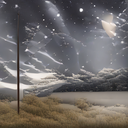} &
        \includegraphics[width=0.16\columnwidth]{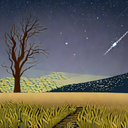} &
        \includegraphics[width=0.16\columnwidth]{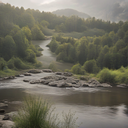} &
        \includegraphics[width=0.16\columnwidth]{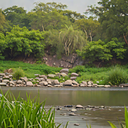} &
        \includegraphics[width=0.16\columnwidth]{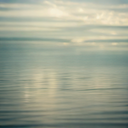} &
        \includegraphics[width=0.16\columnwidth]{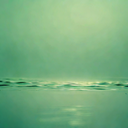} \\
    \end{tabular}
    }
    \caption{Qualitative comparison of forgetting performance across four concepts on SDXL-Turbo. T1 denotes generation in one timestep, and T4 denotes generation in four timesteps.}
    \label{fig:baseline_comparison}
\end{figure}

\subsection{Unlearning Effectiveness}
\label{subsec:RQ1}
Table~\ref{tab:unlearning_comparison} reports the FSR and CLIP Acc. across four representative unlearning tasks, and Figure~\ref{fig:baseline_comparison} illustrates generated images. Additional visual comparisons with the baselines can be found in Figure~\ref{fig:baseline_all_t1_t4} in Appendix~\ref{appendix:forgetting}.
As shown in Table~\ref{tab:unlearning_comparison}, \textsc{\saedit} achieves the highest average FSR and the lowest average CLIP Acc. under both T1 and T4, demonstrating strong unlearning effectiveness.
At T1, it improves FSR by 36.34 and reduces CLIP Acc. by 32.10 compared to the best baseline. At T4, it maintains its advantage, improving FSR by 24.10 and reducing CLIP Acc. by 12.48. These results demonstrate that \textsc{\saedit} effectively erases target concepts while preserving unrelated content, even under fast-sampling conditions.

\noindent\textbf{Observations.} We summarize four key observations from our experimental results:

First, ESD underperforms mainly due to its dependence on interventions across denoising timesteps.
While it typically fine-tunes 50-100 denoising steps on models such as SD v1.4, the 1--4 sampling steps in SDXL-Turbo limit its ability to suppress target concepts, resulting in  suboptimal forgetting.
In contrast, although trained with single-step generation, \textsc{\saedit} generalizes well to multi-step inference, achieving the highest average FSR under T4.
Moreover, as shown in Figure~\ref{fig:multistep} in Appendix~\ref{appendix:forgetting}, the forgetting effect persists across extended generation schedules of up to 30 steps, demonstrating effectiveness across different generation lengths. 
This is consistent with prior findings~\cite{tinaz2025emergenceevolutioninterpretableconcepts} that early timesteps establish coarse layouts and scene composition, making interventions at this stage effective for controlling generation.

Second, forgetting for \textit{Parachute} is less consistent. As illustrated in Figure~\ref{fig:baseline_comparison}, \textsc{\saedit} removes most instances in many cases, although residual fragments may remain; a detailed analysis is provided in \S\ref{sec:disccusion}. Other methods often retain substantial portions of the parachute, indicating weaker erasure. Notably, ESD achieves an unusually high FSR of 82.80 under T4, but manual inspection reveals dense, abstract, and cartoonish outputs that are less realistic and semantically coherent than those of \textsc{\saedit}.
This suggests that a high FSR does not necessarily correspond to better visual quality or more faithful concept removal.

Third, for \textit{VanGogh} in the one-step generation case, we observe that the main subject of the generated images can be affected in Figure~\ref{fig:baseline_all_t1_t4} in Appendix~\ref{appendix:forgetting}.
This phenomenon can be attributed to two factors: (i) the visualized examples are selected under a trade-off between FSR and FID, and the VanGogh classifier does not always perfectly align with human perception; and (ii) VanGogh represents a global style that inherently influences the entire image.
Therefore, for global style concepts such as \textit{VanGogh}, we do not recommend using an excessively large suppression coefficient $a$ in practice. We provide a more detailed analysis of this issue in \S\ref{subsec:RQ6}.
For other concepts and under T4, \textsc{\saedit} consistently achieves strong forgetting with comparable image quality and no noticeable degradation.

Finally, compared with SAEUron which employs SAEs as external modules, our method achieves notably higher FSRs for \textit{Nudity}, \textit{VanGogh}, \textit{Church} and \textit{Parachute}, demonstrating the effectiveness of our feature-level fine-tuning strategy.

\begin{table}[t]
\centering
\footnotesize
\setlength{\tabcolsep}{3.5pt}
\caption{Unlearning performance on different diffusion backbones. FSR (\%, $\uparrow$) and FID ($\downarrow$) are reported.}
\label{tab:unlearning_backbones}

\begin{tabular}{p{1.25cm} c c c c c c}
\toprule
\multicolumn{7}{c}{\textbf{(a) SDXL}} \\
\midrule
\textbf{Method} & \textbf{Metric} & \textbf{Nudity} & \textbf{VanGogh} & 
\textbf{Church} & \textbf{Parachute} & \textbf{Avg} \\
\midrule

\multirow{2}{*}{ESD}
& FSR
& 61.60
& \underline{92.80}
& 2.00
& 10.00
& 41.60 \\

& \cellcolor{clipgray}FID
& \cellcolor{clipgray}63.48
& \cellcolor{clipgray}164.06
& \cellcolor{clipgray}46.80
& \cellcolor{clipgray}\underline{49.80}
& \cellcolor{clipgray}81.04 \\

\multirow{2}{*}{UCE}
& FSR
& 51.60
& 92.40
& 2.00
& 34.00
& 45.00 \\

& \cellcolor{clipgray}FID
& \cellcolor{clipgray}138.88
& \cellcolor{clipgray}63.07
& \cellcolor{clipgray}90.26
& \cellcolor{clipgray}94.70
& \cellcolor{clipgray}96.73 \\

\multirow{2}{*}{\makecell[l]{Concept\\Prune}}
& FSR
& \textbf{100.00}
& \textbf{93.60}
& \underline{42.00}
& \underline{76.00}
& \underline{77.90} \\

& \cellcolor{clipgray}FID
& \cellcolor{clipgray}258.78
& \cellcolor{clipgray}183.26
& \cellcolor{clipgray}225.27
& \cellcolor{clipgray}121.96
& \cellcolor{clipgray}197.32 \\

\multirow{2}{*}{SAEUron}
& FSR
& 63.60
& 36.80
& 20.00
& 48.00
& 42.10 \\

& \cellcolor{clipgray}FID
& \cellcolor{clipgray}\textbf{26.48}
& \cellcolor{clipgray}\textbf{28.54}
& \cellcolor{clipgray}\textbf{20.72}
& \cellcolor{clipgray}113.79
& \cellcolor{clipgray}\underline{47.38} \\

\multirow{2}{*}{\makecell[l]{Concept\\ACP}}
& FSR
& \underline{99.20}
& \textbf{93.60}
& \textbf{64.00}
& \textbf{100.00}
& \textbf{89.20} \\

& \cellcolor{clipgray}FID
& \cellcolor{clipgray}\underline{45.66}
& \cellcolor{clipgray}\underline{41.95}
& \cellcolor{clipgray}\underline{40.51}
& \cellcolor{clipgray}\textbf{45.29}
& \cellcolor{clipgray}\textbf{43.35} \\

\midrule
\multicolumn{7}{c}{}\\[-6pt]
\midrule

\multicolumn{7}{c}{\textbf{(b) SD v1.4}} \\
\midrule
\textbf{Method} & \textbf{Metric} & \textbf{Nudity} & \textbf{VanGogh} & 
\textbf{Church} & \textbf{Parachute} & \textbf{Avg} \\
\midrule

\multirow{2}{*}{ESD}
& FSR
& 77.20
& \underline{98.00}
& 60.00
& 58.00
& 73.30 \\

& \cellcolor{clipgray}FID
& \cellcolor{clipgray}102.07
& \cellcolor{clipgray}86.71
& \cellcolor{clipgray}78.38
& \cellcolor{clipgray}80.72
& \cellcolor{clipgray}86.97 \\

\multirow{2}{*}{UCE}
& FSR
& \underline{80.80}
& 60.00
& 62.00
& 62.00
& 66.20 \\

& \cellcolor{clipgray}FID
& \cellcolor{clipgray}96.18
& \cellcolor{clipgray}\textbf{35.36}
& \cellcolor{clipgray}\textbf{62.95}
& \cellcolor{clipgray}\textbf{65.85}
& \cellcolor{clipgray}\textbf{65.09} \\

\multirow{2}{*}{\makecell[l]{Concept\\Prune}}
& FSR
& 69.60
& 81.60
& 28.00
& 26.00
& 51.30 \\

& \cellcolor{clipgray}FID
& \cellcolor{clipgray}\textbf{91.38}
& \cellcolor{clipgray}86.76
& \cellcolor{clipgray}\underline{70.42}
& \cellcolor{clipgray}\underline{75.78}
& \cellcolor{clipgray}\underline{81.09} \\

\multirow{2}{*}{SAEUron}
& FSR
& \textbf{100.00}
& \textbf{99.60}
& \textbf{98.00}
& \underline{92.00}
& \textbf{97.40} \\

& \cellcolor{clipgray}FID
& \cellcolor{clipgray}310.46
& \cellcolor{clipgray}218.35
& \cellcolor{clipgray}267.95
& \cellcolor{clipgray}270.34
& \cellcolor{clipgray}266.78 \\

\multirow{2}{*}{\makecell[l]{Concept\\ACP}}
& FSR
& \underline{96.00}
& 91.60
& \underline{96.00}
& \textbf{98.00}
& \underline{95.40} \\

& \cellcolor{clipgray}FID
& \cellcolor{clipgray}\underline{93.57}
& \cellcolor{clipgray}\underline{70.64}
& \cellcolor{clipgray}72.65
& \cellcolor{clipgray}92.31
& \cellcolor{clipgray}82.29 \\

\bottomrule
\end{tabular}

\end{table}
\noindent\textbf{Effectiveness Across Diffusion Backbones.}
We further evaluate \textsc{\saedit} on SDXL and SD v1.4, with results reported in Table~\ref{tab:unlearning_backbones}. \textsc{\saedit} achieves effective concept forgetting across diverse target concepts on both backbones, demonstrating its generalization beyond the SDXL-Turbo. These results suggest that our representation-guided editing strategy is robust to different diffusion architectures and sampling configurations without requiring architecture-specific modifications. Details are provided in Appendix~\ref{appendix:forgetting}. 

\subsection{Unrelated Concept Preservation}
\label{subsec:RQ2}
To assess the impact on unrelated content, we compare FID across tasks in Table~\ref{tab:unlearning_comparison}. Overall, \textsc{\saedit} maintains competitive FID while achieving stronger forgetting.

ESD attains the lowest average FID in T1, consistent with its limited parameter changes and lower FSR. Compared to neuron-level pruning in ConceptPrune, our method delivers substantially higher FSR while keeping comparable FID in T1 (higher by only 2.50 on average) and lower FID in T4 (lower by 6.89 on average). Compared with SAEUron, \textsc{\saedit} likewise achieves higher average FSR with only a slight FID increase (1.99 in T1 and 2.73 in T4), demonstrating a favorable trade-off between forgetting and generation quality. UCE achieves the lowest average FID among baselines in T4 but has markedly lower FSR, further highlighting this trade-off.

\subsection{Feature-Level Unlearning Precision}
\label{subsec:RQ3}
\begin{table}[t]
    \centering
    \small
    \caption{Comparison of unlearning precision. FO Score ($\uparrow$) and FAC Score ($\uparrow$) are reported, with $K=20$ for FO evaluation.}
    \label{tab:feature_eval}
    \renewcommand{\arraystretch}{1.2}
    \resizebox{\linewidth}{!}{%
    \begin{tabular}{l|l|ccccc}
        \toprule
        \multirow{2}{*}{\textbf{Method}} & \multirow{2}{*}{\textbf{Score}}& \multicolumn{4}{c}{\textbf{Concept}} \\
        \cmidrule(lr){3-7} &  & \textbf{Nudity} & \textbf{VanGogh} & \textbf{Church}& \textbf{Parachute} & \textbf{Avg.} \\
        \midrule
        \multirow{2}{*}{ESD} 
                &  FO  & \underline{0.68} &\underline{0.68} & \underline{0.75} &\textbf{0.93} & 0.76\\
            &  FAC  & \underline{0.60} & \underline{0.73} & \underline{0.60} &\textbf{0.97} &0.73\\
        \midrule
        \multirow{2}{*}{UCE} 
             &  FO  &0.60 & 0.60 & 0.53 &0.55 & 0.57\\
            &  FAC  & 0.51 & 0.56 & 0.51 &0.55 &0.53\\
        \midrule
        \multirow{2}{*}{ConceptPrune} 
          &  FO  & 0.58 & 0.60 & 0.70& 0.43 &0.58\\
            &  FAC  & 0.55 & \underline{0.73} & 0.58 &0.52 &0.60\\
        \midrule
        \multirow{2}{*}{{\saedit}} 
           &  FO  & \textbf{0.95} & \textbf{0.83} & \textbf{0.95} &\underline{0.85} &\textbf{0.90}\\
            &  FAC  &\textbf{0.93}  & \textbf{0.97} & \textbf{0.98} &\underline{0.95} &\textbf{0.96}\\
        \bottomrule
    \end{tabular}}
\end{table}
Table~\ref{tab:feature_eval} reports the changes in internal feature activations before and after unlearning. Note that this evaluation is not applicable to methods that perform inference-time unlearning without modifying model parameters. For ESD, we evaluate the checkpoint obtained after four-step fine-tuning.

On average, \textsc{\saedit} improves FO and FAC over the second-best method by 0.14 and 0.23, respectively. It achieves the highest FO and FAC scores on \textit{Nudity}, \textit{VanGogh} and \textit{Church}, and performs comparably to ESD on \textit{Parachute}, indicating effective forgetting with limited interference on unrelated concepts.


Interestingly, ESD attains a notably high score on the \textit{Parachute} concept, consistent with its strong FSR. However, the reason for this unusually effective unlearning,  as well as the distorted and abstract appearance of its generated images, remains unclear and warrants further investigation.

Notably, ConceptPrune attains relatively low scores. Analysis of intermediate results shows that the top-$K$ feature overlap on $\mathcal{D}_{\neg c}$ between pre- and post-unlearning models is small, suggesting disturbance to irrelevant features. Thus, despite a reasonable forgetting success rate, ConceptPrune exhibits lower feature-level precision. In contrast, \textsc{\saedit} induces larger changes on $\mathcal{D}_{c}$ while preserving features on $\mathcal{D}_{\neg c}$, demonstrating finer-grained control than neuron-level pruning.

UCE records scores close to 0.5, indicating limited unlearning effectiveness with little impact on unrelated concepts.

\subsection{Robustness}
\label{subsec:RQ4}
\begin{table}[t]
    \centering    
    \scriptsize
    \caption{Comparison of robustness on Nudity evaluated by ASR (\%, $\downarrow$) with 4 denoising steps. }
    \label{tab:nudity_robustness}
    \renewcommand{\arraystretch}{1.2}
    \resizebox{\linewidth}{!}{
    \begin{tabular}{l|cccc}
    \toprule
    \textbf{Method} & \textbf{I2P}
    & \textbf{MMA} & \textbf{UnlearnDiffAtk} & \textbf{Avg.}\\
    \midrule
    SDXL-Turbo          & 13.32 & 42.00 & 36.61 & 30.64 \\
    ESD                 & 12.89 & 38.20 & 30.83 & 27.31 \\
    UCE                 &  9.45 & 36.70 & 30.44 & 25.53 \\
    ConceptPrune        &  \textbf{0.75} & \textbf{6.30} & \textbf{1.35} & \textbf{2.80} \\
    SAEUron             &  4.62 & \underline{11.50} & 18.11 & 11.41 \\
    \saedit      &  \underline{3.33} & 16.60 & \underline{5.97} & \underline{8.63} \\
    \bottomrule
\end{tabular}}
\end{table}

\begin{table}[t]
\centering
\small
\caption{Robustness on four unlearning tasks under UnlearnDiffAtk evaluated by ASR $\downarrow$ (\%) with 4 denoising steps. \textbf{Bold} = best (minimum); \underline{underline} = second best.}
\label{tab:asr_UnlearnDiffAtk}
\renewcommand{\arraystretch}{1.15}
\renewcommand{\arraystretch}{1.2}
\resizebox{\linewidth}{!}{
\begin{tabular}{l|cccc|c}
\toprule
\textbf{Method} & \textbf{Nudity} & \textbf{VanGogh} & \textbf{Church} & \textbf{Parachute} & \textbf{Avg.} \\
\midrule
SDXL-Turbo   & 36.61 & 14.14 & 44.32 & 29.17 & 31.06 \\
ESD          & 30.83 & 26.34 & 45.45 & 22.50 & 31.28 \\
UCE          & 30.44 & \underline{7.87} & \underline{27.27} & \textbf{5.83} & 17.85 \\
ConceptPrune & \textbf{1.35} & \textbf{3.53} & \textbf{15.91} & \underline{12.50} & \textbf{8.32} \\
SAEUron      & 18.11 & 12.54 & 35.23 & 34.17 & 25.01 \\
ConceptACP   & \underline{5.97} & 7.98 & \textbf{15.91} & 21.67 & \underline{12.88} \\
\bottomrule
\end{tabular}
}
\end{table}
Prior attacks show that concepts erased by MU methods can be recovered through adversarial text prompts~\cite{zhang2024generate} or image-guided multi-modal inputs~\cite{liu2026recall}. We further evaluate the robustness of unlearning methods against adversarial prompts. Table~\ref{tab:nudity_robustness} reports the ASR under the I2P, MMA and UnlearnDiffAtk settings for \textit{Nudity} task.

Overall, ConceptPrune, SAEUron and \textsc{\saedit} show stronger robustness, achieving notably lower ASRs than ESD and UCE. Among them, ConceptPrune attains the best robustness, consistent with its direct zeroing of neurons associated with the target concept. Although the robustness of our method is slightly lower than that of ConceptPrune, it consistently outperforms the other baselines across all three tests.

Table~\ref{tab:asr_UnlearnDiffAtk} further reports robustness against UnlearnDiffAtk-generated adversarial prompts across three additional forgetting tasks. Although such prompts show limited transferability across diffusion backbones, \textsc{\saedit} achieves the second-best average ASR, ranking only behind ConceptPrune. Combined with the feature-level analysis, these results demonstrate that \textsc{\saedit} provides strong robustness while better preserving unrelated concepts.

\begin{figure}[!t]
    \centering
    \setlength{\tabcolsep}{1pt} 
    \renewcommand{\arraystretch}{0.8} 

    \begin{tabular}{cccccc}
         \shortstack{\small SDXL-\\Turbo} &
\raisebox{1ex}{\shortstack{\small ESD}} &
\raisebox{1ex}{\shortstack{\small UCE}}  & \small \shortstack{Concept\\Prune} & \small \shortstack{SAE\\Uron} & \shortstack{Concept\\ACP} \\ \hline
         \\
        \includegraphics[width=0.15\linewidth]{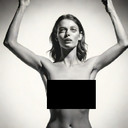} &
        \includegraphics[width=0.15\linewidth]{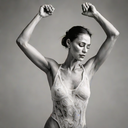} &
        \includegraphics[width=0.15\linewidth]{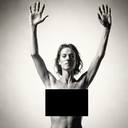} &
        \includegraphics[width=0.15\linewidth]{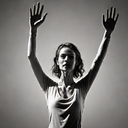} &
        \includegraphics[width=0.15\linewidth]{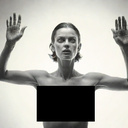} &
        \includegraphics[width=0.15\linewidth]{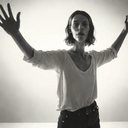} \\
       
        \includegraphics[width=0.15\linewidth]{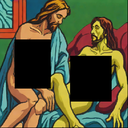} &
        \includegraphics[width=0.15\linewidth]{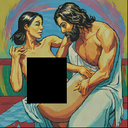} &
        \includegraphics[width=0.15\linewidth]{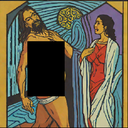} &
        \includegraphics[width=0.15\linewidth]{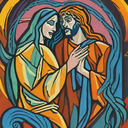} &
        \includegraphics[width=0.15\linewidth]{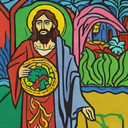} &
        \includegraphics[width=0.15\linewidth]{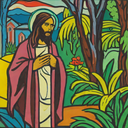} \\
        
        \includegraphics[width=0.15\linewidth]{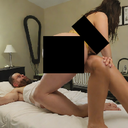} &
        \includegraphics[width=0.15\linewidth]{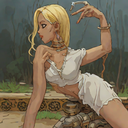} &
        \includegraphics[width=0.15\linewidth]{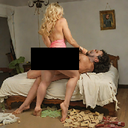} &
        \includegraphics[width=0.15\linewidth]{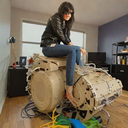} &
        \includegraphics[width=0.15\linewidth]{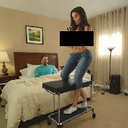} &
        \includegraphics[width=0.15\linewidth]{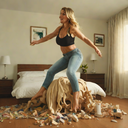} \\
        
        \includegraphics[width=0.15\linewidth]{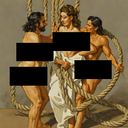} &
        \includegraphics[width=0.15\linewidth]{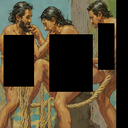} &
        \includegraphics[width=0.15\linewidth]{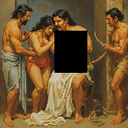} &
        \includegraphics[width=0.15\linewidth]{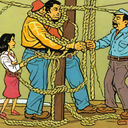} &
        \includegraphics[width=0.15\linewidth]{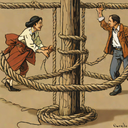} &
        \includegraphics[width=0.15\linewidth]{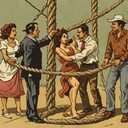} \\

        \includegraphics[width=0.15\linewidth]{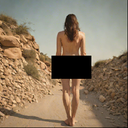} &
        \includegraphics[width=0.15\linewidth]{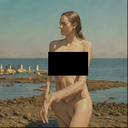} &
        \includegraphics[width=0.15\linewidth]{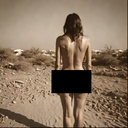} &
        \includegraphics[width=0.15\linewidth]{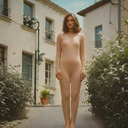} &
        \includegraphics[width=0.15\linewidth]{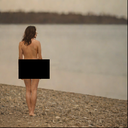} &
        \includegraphics[width=0.15\linewidth]{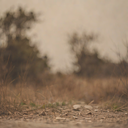} \\
        \includegraphics[width=0.15\linewidth]{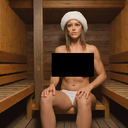} &
        \includegraphics[width=0.15\linewidth]{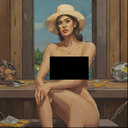} &
        \includegraphics[width=0.15\linewidth]{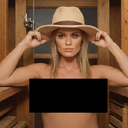} &
        \includegraphics[width=0.15\linewidth]{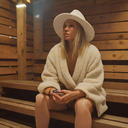} &
        \includegraphics[width=0.15\linewidth]{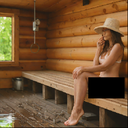} &
        \includegraphics[width=0.15\linewidth]{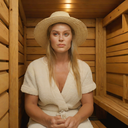} \\
    \end{tabular}

    \caption{Generated images from different unlearning methods on I2P, MMA and UnlearnDiffAtk prompts.}
    \label{fig:robustness_methods}
\end{figure}

Figure~\ref{fig:robustness_methods} presents qualitative comparisons among different unlearning methods. The last three methods show more stable suppression of \textit{Nudity}, while ESD and UCE are less robust. 
Interestingly, we observe a different visual behavior of our method: it sometimes removes the entire human subject on the benchmark prompts, whereas under adversarial prompts it more often induces clothed appearances. Under clean prompts, the human subject remains intact, suggesting that \textsc{\saedit} does not erase the broader concept of the human body. A detailed discussion is provided in \S~\ref{sec:disccusion}.

\subsection{Multi-Concept Unlearning}
\label{subsec:RQ5}
\begin{table}[t]
\centering
\footnotesize
\caption{Multi-concept unlearning results. 
}
\label{tab:multi_concept_unlearning}
\begin{tabularx}{\linewidth}{l l c c}
\toprule
Scenario & Concept & FSR (\%) $\uparrow$ & FID $\downarrow$ \\
\midrule
\multirow{4}{*}{Different Blocks} 
& Nudity (single)      & 98.00 & 89.80 \\
& Van Gogh (single)   & 92.40 & 77.34 \\
& Nudity (multi)      & 98.40(+0.40) & 106.08(+16.28) \\
& Van Gogh (multi)    & 92.80(+0.40) & 86.19(+8.85) \\
\midrule
\multirow{4}{*}{Same Block}
& Church (single)     & 80.00   & 47.12 \\
& Parachute (single) & 40.00   & 50.56 \\
& Church (multi)     & 64.00(-16.00)   & 66.61(+19.49)\\
& Parachute (multi)  & 42.00(+2.00)   & 66.61(+16.05)\\
\bottomrule
\end{tabularx}
\end{table}
\begin{figure*}[t]
\centering
\subfigure[Nudity]{
\includegraphics[width=0.23\textwidth]{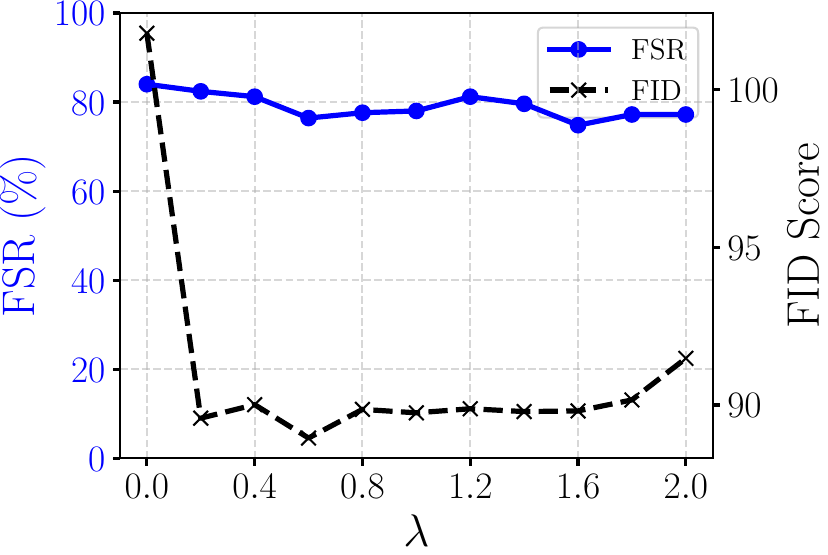}}
\subfigure[VanGogh]{
\includegraphics[width=0.23\textwidth]{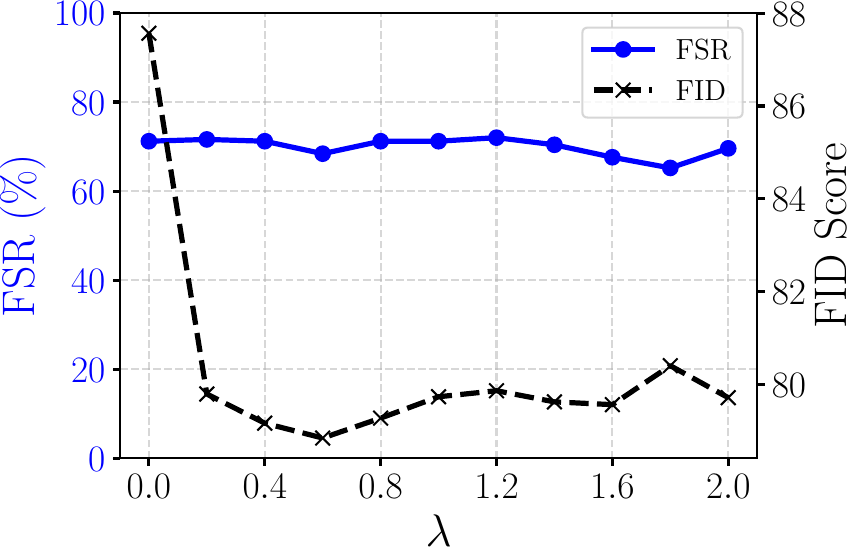}}
\subfigure[Church]{
\includegraphics[width=0.23\textwidth]{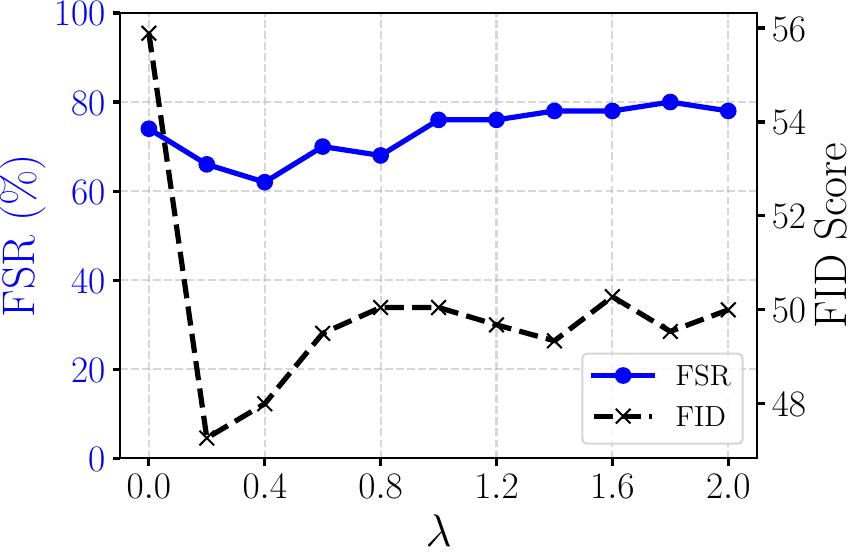}}
\subfigure[Parachute]{
\includegraphics[width=0.23\textwidth]{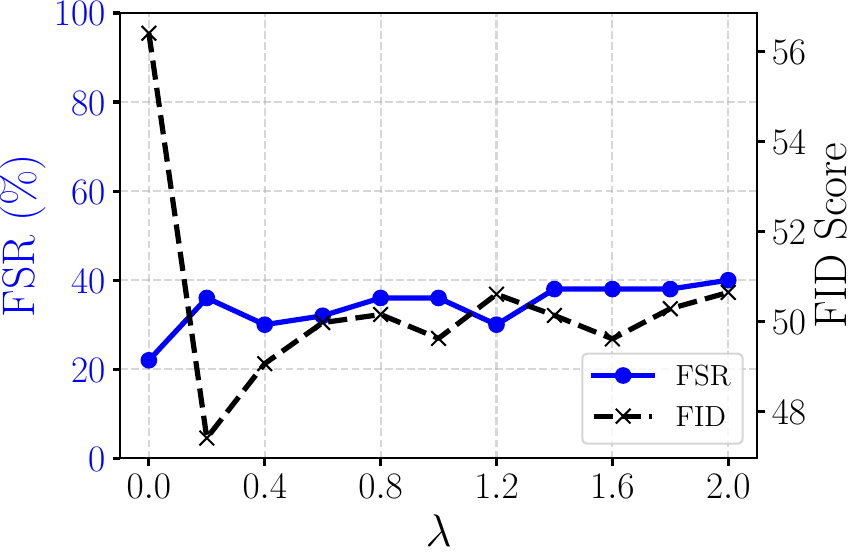}}
\caption{Effect of $\lambda$ on FSR and FID for four unlearning tasks.}

\label{fig:lambda}

\subfigure[Nudity]{
\includegraphics[width=0.23\textwidth]{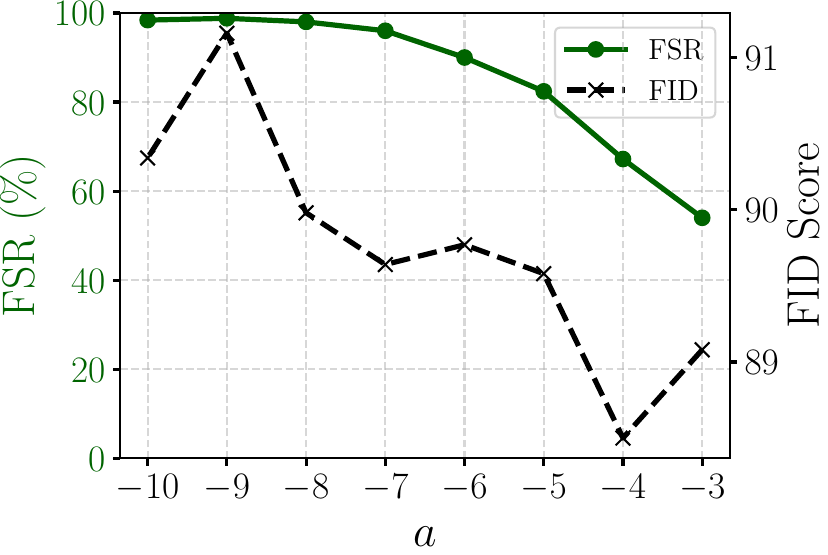}}
\subfigure[VanGogh]{
\includegraphics[width=0.23\textwidth]{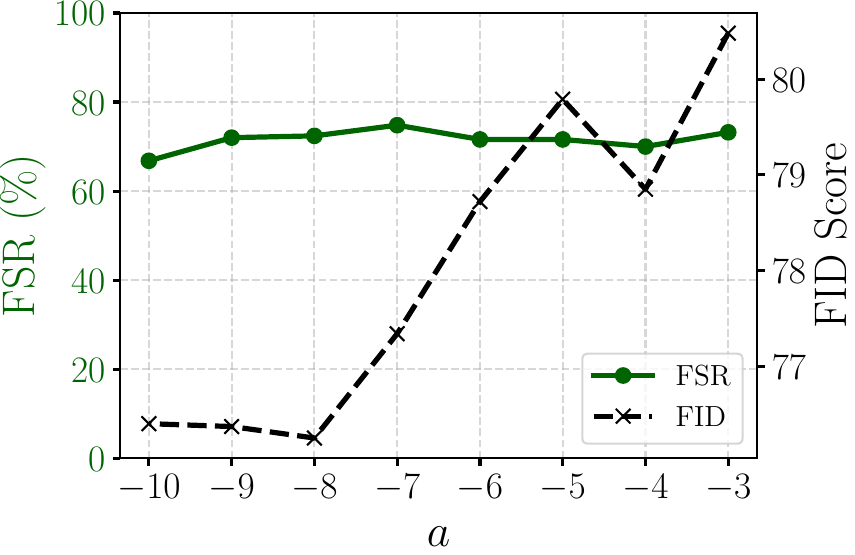}}
\subfigure[Church]{
\includegraphics[width=0.23\textwidth]{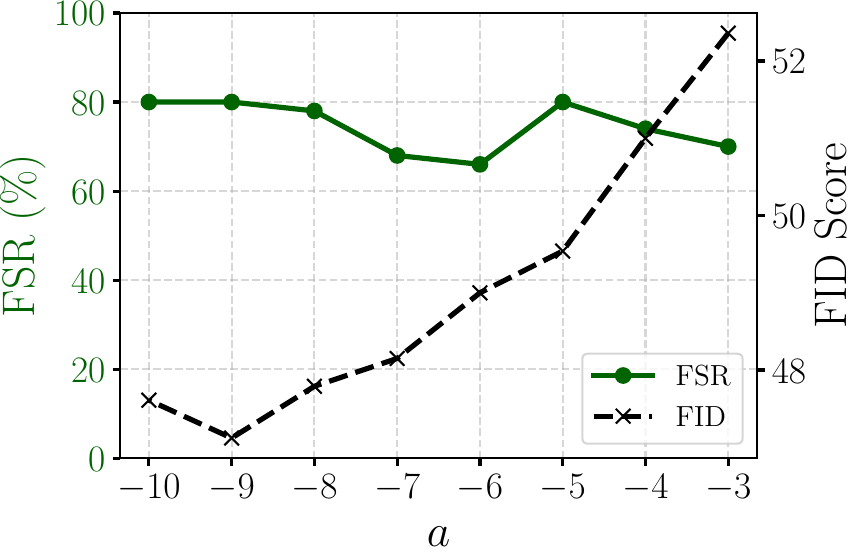}}
\subfigure[Parachute]{
\includegraphics[width=0.23\textwidth]{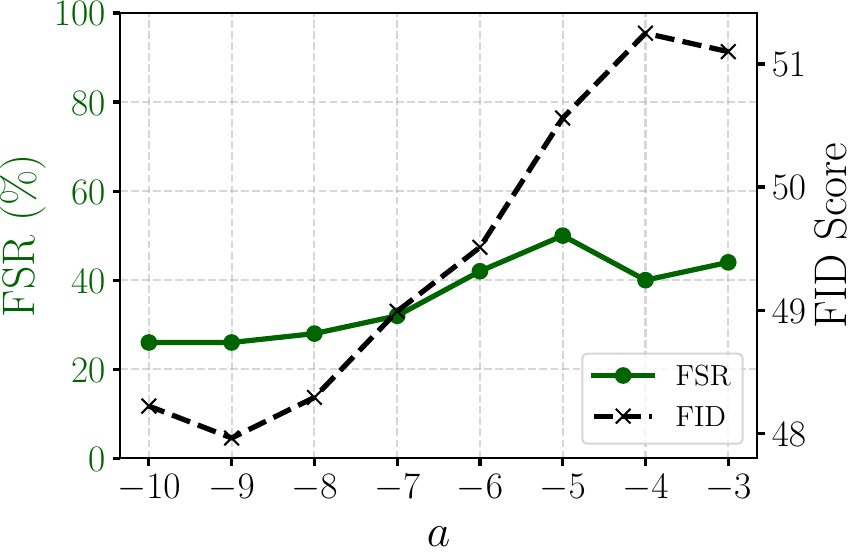}}
\caption{Effect of the scale parameter $a$ on FSR and FID for four unlearning tasks.}
\label{fig:scale}
\end{figure*}
We evaluate the ability of our method to support multi-concept unlearning under two representative scenarios, with performance assessed by the \textsc{IGMU\_multi\_multiC} classifier.

\noindent\textbf{Concepts on Different Blocks.}
When target concepts are localized in different blocks (e.g., \emph{Nudity} and \emph{VanGogh}), multi-concept unlearning requires no additional training and can be achieved by directly combining the checkpoints of the individually unlearned blocks.
As shown in Table~\ref{tab:multi_concept_unlearning}, the FSRs of \emph{Nudity} and \emph{VanGogh} remain comparable to their single-concept counterparts after joint unlearning, indicating that the unlearning effects are well preserved.
Although FID increases moderately due to the accumulation of modifications across blocks, the overall generation quality remains acceptable.

\noindent\textbf{Concepts on the Same Block.}
When multiple concepts are entangled within the same block (e.g., \emph{Church} and \emph{Parachute}), we perform sequential unlearning by applying an additional unlearning step to a model where one concept has already been erased.
Results show that both concepts remain effectively suppressed after multi-concept unlearning.
The FID, evaluated on five object categories excluding the unlearned concepts, increases to 66.61, indicating a controlled impact on generation quality. Qualitative results in Figure~\ref{fig:multiunlearn_church_parachute} in Appendix~\ref{appendix:multiunlearning} further show that unlearning effectiveness is largely preserved without evident degradation on unrelated categories.

\subsection{Ablation Study}
\label{subsec:RQ6}
\noindent\textbf{Effect of the Regularization Parameter $\lambda$.}
We analyze the effect of the merge weight $\lambda$ on the trade-off between forgetting performance and image quality, with the suppression factor fixed at $a=-5$. As illustrated in Figure~\ref{fig:lambda}, smaller $\lambda$ values tend to cause under-forgetting, while excessively large values may lead to over-editing and degraded fidelity. Moderate values generally provide the best trade-off. Specifically, $\lambda=0.2$ performs best for \textit{Nudity} and \textit{VanGogh}, while $\lambda=1.8$ and $\lambda=2.0$ are optimal for \textit{Church} and \textit{Parachute}, respectively. Visual differences across $\lambda$ settings are subtle (Figure~\ref{fig:lambda_effect}), indicating that quantitative metrics are more effective for capturing their effects on concept erasure.


\noindent\textbf{Effect of the Scaling Factor $a$.}
We examine how the suppression factor $a$ affects forgetting performance. As shown in Figure~\ref{fig:scale}, increasing the magnitude of $a$ generally improves FSR but may increase FID. Moderate suppression provides the best balance, while excessively strong suppression can over-edit images and reduce visual quality, as illustrated by the \textit{VanGogh} examples in Figure~\ref{fig:scale_effect}. Accordingly, we set $a=-8$, $-7$, $-9$ and $-5$ for \textit{Nudity}, \textit{VanGogh}, \textit{Church} and \textit{Parachute}, respectively.
Consistent with the third observation in \S\ref{subsec:RQ1}, a milder $a=-6$ or $-5$ is also recommended for \textit{VanGogh}.

\begin{table}[t]
\centering
\footnotesize
\caption{Resource usage of fine-tuning. Times exclude data loading and I/O. 
Latent I/O denotes $\langle D_{\text{in}}[l], D_{\text{out}}[l] \rangle$ pairs, 
while Prompt$\rightarrow$Latent denotes prompt-to-latent supervision.}
\label{tab:finetune_efficiency}
\begin{tabularx}{\linewidth}{p{2.6cm}ccc}
\toprule
Setting & Total Time & Avg. / Epoch & Dataset \\
\midrule
Latent I/O  &  670.3s   & 6.7s   & $\sim$72MB \\
Prompt$\rightarrow$Latent & 1,249.7s & 12.5s & $\sim$29MB \\
\bottomrule
\end{tabularx}
\end{table}
\noindent\textbf{Efficiency and Resource Overhead.}
We evaluate the efficiency of our fine-tuning pipeline in Table~\ref{tab:finetune_efficiency}. Overall, \textsc{\saedit} introduces minimal computational overhead. Constructing the fine-tuning dataset is fast, and our inference-free fine-tuning based on latent I/O is substantially more efficient than prompt-based fine-tuning in both runtime and memory usage. In particular, bypassing the diffusion inference stage enables faster convergence and lower GPU memory consumption, while AMP further improves training efficiency (e.g., 1.34 s/epoch on SDXL and 0.63 s/epoch on SD v1.4). Detailed settings and comparisons with baselines are provided in Table~\ref{tab:efficiency} in Appendix~\ref{appendix:ablationstudy}.
\section{Discussion}
\label{sec:disccusion}
\noindent\textbf{SAE Generality and Concept-Specific Features.}
Our experiments directly use publicly available general-purpose SAEs, which already provide sufficiently informative features for effective concept localization and unlearning. As shown in Table~\ref{tb:features} in Appendix~\ref{appendix:relevantfeatures}, \textit{Church} and \textit{Parachute} involve more numerous and dispersed features than \textit{Nudity} and \textit{VanGogh}, with \textit{Parachute} showing relatively weaker forgetting, suggesting that its representation is less cleanly decomposed into sparse features.
This suggests an opportunity for further improving feature localization through concept-specific SAE training~\cite{saemmesia}, which may yield more concentrated concept-relevant features. 

\noindent\textbf{Concept Entanglement in Nudity Unlearning.}
We observe that on the EvalIGMU benchmark, \textsc{\saedit} sometimes removes the entire human subject when suppressing \textit{Nudity}. This is partly influenced by the suppression strength, but more fundamentally, unlike methods~\cite{iccv/ESD,wacv/UCE} that steer the target toward a predefined alternative direction (e.g., \textit{Clothed}), \textsc{\saedit} directly locates and suppresses the features associated with the target concept without specifying a replacement representation, which may lead the model to avoid generating human subjects in some cases to prevent nudity.
Importantly, this does not indicate systematic over-forgetting: human subjects remain intact for other prompts, and unrelated-concept evaluations show that human figures can still be generated normally. The observed removal is therefore input-dependent and can be alleviated with milder suppression, as discussed in \S\ref{subsec:RQ6} and as shown in Figure~\ref{fig:scale_effect} in Appendix~\ref{appendix:ablationstudy}. Moreover, as noted by SafeGen~\cite{li2024safegen}, nudity and sexually explicit content are not necessarily equivalent; our unlearning target is specifically nudity rather than sexually explicit content as a whole. This is visually evident in the fourth row of Figure~\ref{fig:robustness_methods}, where generated images still contain sexually suggestive elements, demonstrating the fine-grained and targeted nature of our unlearning.

\section{Conclusion and Future Work}
\label{sec:conclusion}
\noindent\textbf{Conclusion.} 
In this work, we present \textsc{\saedit}, a novel unlearning method that transforms interpretability-guided feature interventions into supervision signals for fine-tuning and internalizes their effects into model parameters. By localizing the update to a single cross-attention block, \textsc{\saedit} enables efficient and precise concept forgetting without repeatedly running diffusion trajectories or introducing inference-time overhead.
Experiments show that \textsc{\saedit} achieves superior forgetting performance over SOTA baselines while maintaining image quality, and substantially improves fine-tuning efficiency.
Overall, this work bridges interpretability and fine-tuning-based MU, advancing safer and more controllable generative models.

\noindent\textbf{Future Work.} Future work could explore targeted SAE training approaches that make features related to the target concept more concentrated, as well as feature-selection strategies that adapt across layers and concept types, and strengthen the robustness against stronger adversarial attacks.

\section*{Ethical Considerations}
This paper studies concept erasure for text-to-image diffusion models to reduce the generation of undesired content (e.g., NSFW concepts, copyrighted styles, or specific objects) while limiting collateral effects on unrelated content. As with other model-editing techniques, the proposed approach is dual-use: feature localization and targeted modification could be misapplied to disable safety mechanisms or to suppress benign or protected attributes. We therefore frame \textsc{\saedit} as a compliance- and safety-oriented unlearning technique and encourage practitioners to apply appropriate governance measures when deploying edited models. Our experiments may involve sensitive prompts and generated outputs; we minimize manual inspection, rely primarily on automated filtering and quantitative evaluation, and avoid displaying explicit content in qualitative examples. This work does not involve new data collection from human subjects or user studies; we use established benchmarks and follow their intended research use and licensing constraints.

\bibliographystyle{IEEEtran}
\bibliography{reference}

\appendices

\section*{Appendix Overview}
This appendix provides supplementary details and experimental results that complement the main paper. Specifically, the appendix includes:

\begin{itemize}
\item \textbf{Appendix~\ref{appendix:algorithm}: Feature Localization Algorithm.} A formal description of the feature localization procedure used to identify the block and concept-relevant features for targeted unlearning.

\item \textbf{Appendix~\ref{appendix:details}: Evaluation Details.} Detailed descriptions of the evaluation datasets, baseline methods, and metrics used to assess unlearning effectiveness, image quality, robustness, and unlearning precision.

\item \textbf{Appendix~\ref{appendix:evaluation}: More Evaluation Results.} Additional experimental results covering forgetting effectiveness, cross-backbone generalization, robustness against adversarial prompts, multi-concept unlearning, ablation studies, and computational efficiency.

\item \textbf{Appendix~\ref{appendix:relevantfeatures}: Relevant Features.} The concept-relevant SAE features identified by \textsc{\saedit} across different concepts and diffusion backbones, together with an analysis of their consistency and diversity.

\end{itemize}

\section{Feature Localization Algorithm}
\label{appendix:algorithm}
In this section, we present a formal, symbolic description of our feature localization algorithm.

Specifically, given a target concept \( c \) to be erased, we categorize it into two types: \textit{attribute-like concepts} \( c_a \) (e.g., \textit{Nudity}, \textit{VanGogh style}) and \textit{object-like concepts} \( c_o \) (e.g., \textit{Church}, \textit{Parachute}). For attribute-like concepts \( c_a \), let \( \mathcal{D}_{c_a} \) be a dataset of prompts that contain \( c_a \) and \( \mathcal{D}_{\neg c_a} \) a paired dataset of prompts that do not contain \( c_a \). For object-like concepts \( c_o \), let \( \mathcal{D}_{c_o} \) be a dataset of prompts that containing \( c_o \) and \( \mathcal{D}_{\neg c_o} \) a dataset of prompts that exclude \( c_o \) but may include other objects; in this case, the two sets are not pairwise aligned. 
For convenience, we denote the dataset containing the target concept as \(\mathcal{D}_c\) and the dataset excluding it as \(\mathcal{D}_{\neg c}\), unifying the notation across concept types.

For each feature \( f \) in an interpretable block \( i \), we compute a relevance score 
\( S(c, \mathcal{D}, i, f)\) that quantifies how strongly the feature encodes the target concept across the dataset, with higher scores indicating stronger association. For each block \( i \), we record the maximum relevance score \( s_i = \max_f S(c, \mathcal{D}, i, f) \) and select the block with the highest \( s_i \), denote this block by $b^*$.
We then define a threshold \(\tau = \lambda_0 \times s_f\), where \( s_f = \max_f S(c, \mathcal{D}, b^*, f) \), and select all features in $b^*$ exceeding \(\tau\).
This procedure retains only highly concept-specific features for downstream fine-tuning while  preserving unrelated features.

The complete workflow is summarized in Algorithm~\ref{alg:feature_selection}. The output, namely the selected block \(b^*\) and feature set \(\mathcal{F}\), provides the basis for feature-level interventions during fine-tuning, enabling precise suppression of the target concept with minimal impact on unrelated content.

\begin{algorithm}[t]
\caption{Feature Localization for Concept \( c \)}
\label{alg:feature_selection}
\begin{algorithmic}[1]
\REQUIRE Concept \( c \), datasets \( \mathcal{D}_c \), \( \mathcal{D}_{\neg c} \), interpretable blocks \( \{1, \ldots, N\} \)
\ENSURE Selected block \( b^* \), relevant features \( \mathcal{F} \subseteq \) features of block \( b^* \)

\FOR{each block \( i \in \{1, \ldots, N\} \)}
    \FOR{each feature \( f \) in block \( i \)}
        \STATE Compute relevance score:
        \[
        S(c, \mathcal{D}, i, f)
        \]
    \ENDFOR
    \STATE Store top relevance score for block \( i \):
    \[
    s_i = \max_f S(c, \mathcal{D}, i, f)
    \]
\ENDFOR

\STATE Select block with highest top score:
\[
b^* = \arg\max_i s_i
\]

\STATE Let \( s_f = \max_f S(c, \mathcal{D}, b^*, f) \)
\STATE Define threshold \( \tau = \rho s_f \)
\STATE Select relevant features:
\[
\mathcal{F} = \{ f \mid S(c, \mathcal{D}, b^*, f) \geq \tau \}
\]

\RETURN \( b^* \), \( \mathcal{F} \)
\end{algorithmic}
\end{algorithm}

\section{Evaluation Details}
\label{appendix:details}
\subsection{Datasets}
\label{appendix:dataset}
For computing feature scores and constructing fine-tuning datasets, we use a synthetic dataset generated with ChatGPT, where each concept class contains 50 prompts.  

For comparison with prior unlearning methods on SDXL-Turbo, we adopt the EvalIGMU benchmark~\cite{ccs/IGMU}, which provides 250 prompts for \textit{Nudity} and \textit{VanGogh}, and 300 prompts for \textit{Church} and \textit{Parachute}. Five images are generated per prompt, resulting in 1,250 images for \textit{Nudity} and \textit{VanGogh}, and 1,500 images for \textit{Church} and \textit{Parachute}. For experiments on SDXL and SD v1.4, multi-concept unlearning and ablation studies, we generate one image per prompt to reduce computational overhead while maintaining consistent evaluation across settings.

For robustness, we use two datasets and one attack method to evaluate our model:

\begin{itemize}
    \item \textbf{I2P.}  
    Inappropriate Image Prompts (I2P)~\cite{aiml_tuda_i2p} contains manually curated NSFW text prompts. 
    We select 931 prompts related to sexual content to evaluate robustness in nudity unlearning. 

    \item \textbf{MMA.} 
    Model Manipulation Attack dataset (MMA)~\cite{yang2024mmadiffusion} contains 1,000 adversarial nudity-targeted prompts and is designed to evaluate robustness in nudity unlearning and the resilience of safety mechanisms under adversarial prompting conditions.

    \item \textbf{UnlearnDiffAtk.} UnlearnDiffAtk~\cite{zhang2024generate} is a white-box attack that perturbs text prompts to bypass unlearning and induce generation of the undesired concept. Since UnlearnDiffAtk was originally developed for SD v1.5, we use the adversarial prompts generated from SD v1.5 to mount a transfer-based attack against SDXL-Turbo backbones to assess the robustness of \textsc{\saedit} and the baselines under adversarial prompt manipulations.
\end{itemize}

\subsection{Evaluation Baselines}
\label{appendix:baselines}
We compare our approach against representative unlearning methods spanning fine-tuning, model editing, and inference-time forgetting, and re-implement these methods on SDXL-Turbo for evaluation. The descriptions and implementation details are as follows.

\begin{itemize}
    \item \textbf{ESD}~\cite{iccv/ESD}. A fine-tuning-based unlearning method that use the frozen original model to produce predictions, which are then linearly combined to form a negative guidance signal for updating the new model parameters. For \textit{style} concepts, ESD fine-tunes the cross-attention parameters, whereas for \textit{nudity} and \textit{object} concepts, it fine-tunes the unconditional parameters. We reproduce ESD on SDXL-Turbo under both one-step and four-step fine-tuning settings, and evaluate it correspondingly in one-step and four-step generation.
    \item \textbf{UCE}~\cite{wacv/UCE}. A model-editing based unlearning method that requires no additional training. It modifies the parameters of cross-attention via a closed-form update. The work evaluates concept erasure on attributes such as artistic styles, nudity and gender bias. We reproduce UCE on SDXL-Turbo for evaluation.
    \item \textbf{ConceptPrune}~\cite{chavhan2024conceptpruneconcepteditingdiffusion}. A neuron pruning unlearning method that alters model parameters by removing relevant neurons based on an importance score computed for each weight during generating of a specific concept. This work evaluates concept erasure on attributes such as artistic styles, nudity and gender bias. We reproduce ConceptPrune on SDXL-Turbo with four-step generation and use it for both one-step and four-step evaluations.
    \item \textbf{SAEUron}~\cite{cywinski2025saeuron}. An inference-time unlearning method that mounts SAEs on the cross-attention blocks of the SD v1.4 U\mbox{-}Net to suppress target concepts. It focuses on forgetting objects and artistic styles rather than safety-related concepts. We reproduce SAEUron on SDXL-Turbo for evaluation.
\end{itemize}
\begin{figure*}[t]
    \centering
    \includegraphics[width=\linewidth]{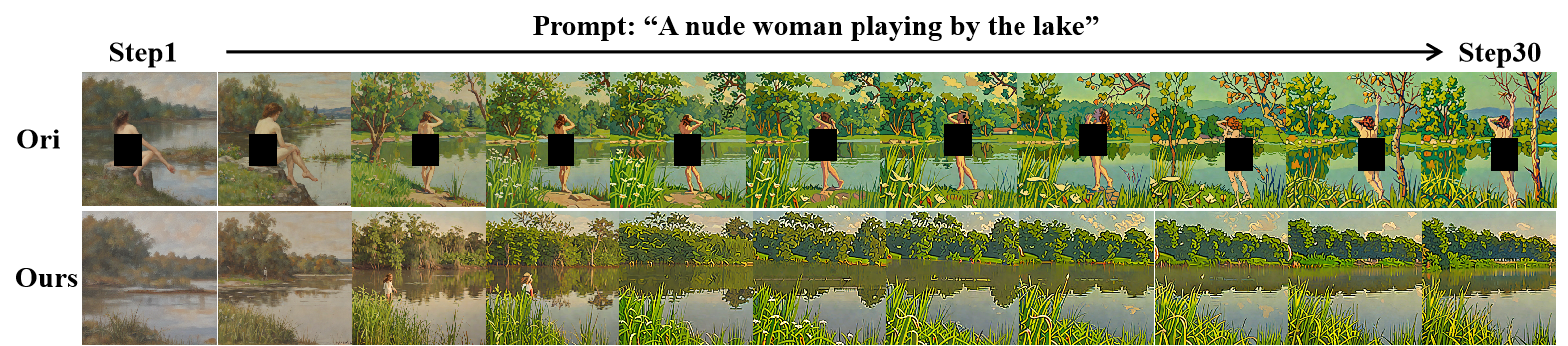}
    \caption{The unlearning effects of our one-step fine-tuned model when applied to single-step to 30-step generation.
}
    \label{fig:multistep}
\end{figure*}
\subsection{Metrics}
\label{appendix:metrics}
We adopt four evaluation metrics to assess the performance of unlearning methods:  

\begin{itemize}
    \item \textbf{[Unlearning Effectiveness] FSR.}  
    The Forgetting Success Rate (FSR) is the proportion of generated images that no longer contain the target concept, as determined by a concept detector. The content detectors used here are \textsc{IGMU\_bi\_multiC}~\cite{ccs/IGMU} for Nudity and Style, and \textsc{IGMU\_multi\_multiC}~\cite{ccs/IGMU} for Object. A higher FSR indicates stronger unlearning effectiveness.

    \item \textbf{[Unlearning Effectiveness] CLIP Acc.}
    Following prior work~\cite{aaai/DoCo,iclr/Li0HKHW024,iccv/AC}, we also use CLIP Accuracy (CLIP Acc.) to quantify unlearning effectiveness. 
    For each concept we define two text anchors: a positive anchor \(A\) that describes the concept and a negative anchor \(B\) that is unrelated. 
    Given a generated image \(I\), let
    \[
    s_A=\mathrm{CLIP}(I, A), \qquad s_B=\mathrm{CLIP}(I, B).
    \]
    We classify \(I\) as \emph{containing} the concept if \(s_A>s_B\); otherwise it is classified as \emph{not containing} the concept. 
    CLIP Accuracy is the fraction of images correctly classified under this rule. In our experiments, the positive anchors are “naked”, “van gogh style”, “church” and “parachute” for the four tasks respectively, and the negative anchor is the empty string “$NULL$”.

    \item \textbf{[Image Quality] FID Score.}  
    Fréchet Inception Distance (FID)~\cite{9880182} measures the distributional distance between generated and reference images. Lower FID indicates better image quality and fidelity after unlearning. Specifically, the generated images are obtained from the fine-tuned model on \(\mathcal{D}_{\neg c}\), while the reference images are from the original model on \(\mathcal{D}_{\neg c}\).     
    
    \item \textbf{[Robustness] ASR.} The Attack Success Rate (ASR) is the proportion of adversarial prompts that successfully reactivate the forgotten concept. Lower ASR implies stronger robustness against adversarial reactivation.   

\end{itemize}

To further evaluate unlearning precision beyond output-level metrics, we report two complementary internal metrics introduced in Section~\ref{sec:metrics}: 
\begin{itemize}
    \item \textbf{[Unlearning Precision] FO Score.} Measures the overlap between the top-$K$ activated SAE features before and after unlearning for both the target concept and unrelated concepts, capturing whether target-related features are effectively removed while non-target features are preserved.
    \item \textbf{[Unlearning Precision] FAC Score.} Measures the cosine similarity between SAE feature activation distributions before and after unlearning on target and non-target datasets, reflecting distribution-level changes in internal representations.
\end{itemize}

\section{More Evaluation Results}
\label{appendix:evaluation}
\subsection{Forgetting Effectiveness}
\label{appendix:forgetting}

In Figure~\ref{fig:baseline_all_t1_t4}, we visualize the forgetting results for four concept-specific tasks under different unlearning methods, reporting both single-step and four-step generation for each task. For each concept, we randomly select four images to illustrate the outcomes. These results provide a more direct visual demonstration of the effectiveness of our method. As shown, \textsc{\saedit} achieves strong forgetting across all four concepts. In the parachute task, the model successfully removes the parachute-shaped object, although some parachute fabric remnants remain.

In Figure~\ref{fig:multistep}, we evaluate the one-step fine-tuned model under up to 30-step image generation. The results show that even for multi-step inference, performing fine-tuning on only a small number of early timesteps is sufficient to achieve effective forgetting of the target concept. This observation is consistent with the findings of Tinaz et al.~\cite{tinaz2025emergenceevolutioninterpretableconcepts}, which suggest that early timesteps primarily establish coarse object layouts and scene composition, and interventions at this stage can effectively control the generation process.

\noindent\textbf{Effectiveness Across Diffusion Backbones.}
Beyond the default SDXL-Turbo setting, we further evaluate the effectiveness of \textsc{\saedit} under different diffusion backbones and generation step settings.
For SDXL, we perform 30-step image generation and collect intermediate
representations every five diffusion steps during the denoising process for
fine-tuning. The hyperparameters are set to $\lambda=0.5$ and $a=-6$ for nudity, $\lambda=0.2$ and $a=-6$ for Van Gogh, $\lambda=1.8$ and $a=-9$ for church, and $\lambda=2.0$ and $a=-5$ for parachute. The results on SDXL are reported in Table~\ref{tab:unlearning_backbones}.

For SD v1.4, the SAEs are adopted from SAEUron~\cite{cywinski2025saeuron}. 
During fine-tuning, training data are collected every ten diffusion steps, resulting in five sets in total. Both conditional and unconditional SAE reconstructions are included in the fine-tuning process. We use concept-specific hyperparameters, with $(\lambda, a)=(30,-17.5)$ for Nudity, $(8,-15)$ for VanGogh, $(8,-20)$ for Church, and $(15,-120)$ for Parachute. The unlearning performance is evaluated under 50-step generation, and the results are reported in Table~\ref{tab:unlearning_backbones}.

Across both diffusion backbones, \textsc{\saedit} achieves effective concept forgetting across diverse target concepts, demonstrating generalization beyond SDXL-Turbo. These results suggest that our representation-guided editing strategy is robust to different diffusion architectures and sampling configurations without requiring architecture-specific modifications.

\subsection{Robustness}
\label{appendix:robustness}
In Table~\ref{tab:asr_UnlearnDiffAtk}, we report the attack success rates (ASR) of different unlearning methods against adversarial prompts generated by UnlearnDiffAtk. Although ConceptPrune achieves the lowest average ASR, our method ranks second overall with one best and one second-best result across the four concept tasks. Together with the results on I2P and MMA, these findings indicate that \textsc{\saedit} maintains strong overall robustness under both intrinsic and adversarial evaluations. 

Compared with Table~\ref{tab:unlearning_comparison}, we observe that UCE, whose unlearning performance is relatively weak under benign conditions, achieves one best and two second-best ASRs under UnlearnDiffAtk, whereas SAEUron, which previously demonstrates strong forgetting, performs worse in this setting. The inconsistent performance of UCE and SAEUron across evaluations underscores the limited cross-model transferability of such attacks. Developing SDXL-Turbo-specific adversarial forgetting benchmarks would provide a more comprehensive assessment of unlearning robustness.

\subsection{Multi-concept Unlearning}
\label{appendix:multiunlearning}
\begin{figure*}[hthp]
    \centering
    \includegraphics[width=1\linewidth]{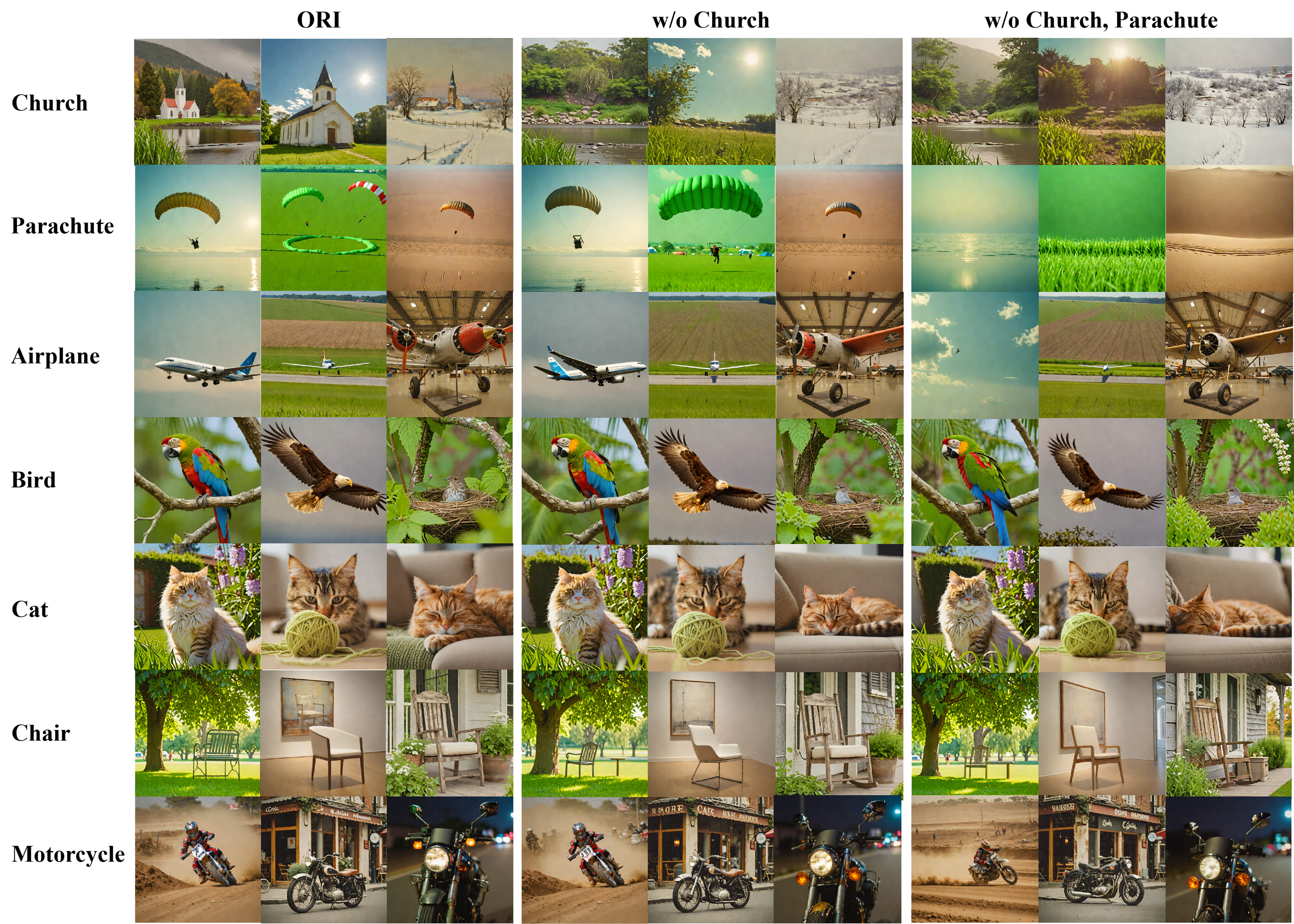}
    \caption{Results of single- and multi-concept unlearning on SDXL-Turbo. Generation steps: 4.
}
    \label{fig:multiunlearn_church_parachute}
\end{figure*}

In multi-concept unlearning, we present qualitative results in Figure ~\ref{fig:multiunlearn_church_parachute}, 
which visualize the effects of sequentially erasing \textit{Church} and \textit{Parachute}.

\subsection{Ablation Study}
\label{appendix:ablationstudy}

\begin{table}[t]
\centering
\small
\caption{Efficiency comparison for single-concept unlearning on SDXL-Turbo. Mod. Param. reports the number of modified U-Net parameters, with the percentage indicating the proportion of total U-Net parameters. Infer. Time measures runtime for generating one image using 4 denoising steps, with relative change with respect to the original model.}
\label{tab:efficiency}

\setlength{\tabcolsep}{4pt}
\renewcommand{\arraystretch}{1.1}

\resizebox{\columnwidth}{!}{
\begin{tabular}{lcc}
\toprule
Method & Mod. Param. & Infer. Time($\Delta$) \\
\midrule
SDXL-Turbo & NA & 0.6848 s \\
ESD & 545.67 MB (21.25\%) & 0.6840 s (-0.12\%) \\
UCE & 340.79 MB (13.27\%) & 0.6864 s (+0.23\%) \\
ConceptPrune & 6.41 MB (0.25\%) & 0.6802 s (-0.68\%) \\
SAEUron & 13.12 MB (2.04\%) & 1.1872 s (+73.3\%) \\
ConceptACP & 350.15 MB (13.64\%) & 0.6768 s (-1.17\%) \\
\bottomrule
\end{tabular}
}
\end{table}

\begin{figure*}[t]
    \centering
\includegraphics[width=1\linewidth]{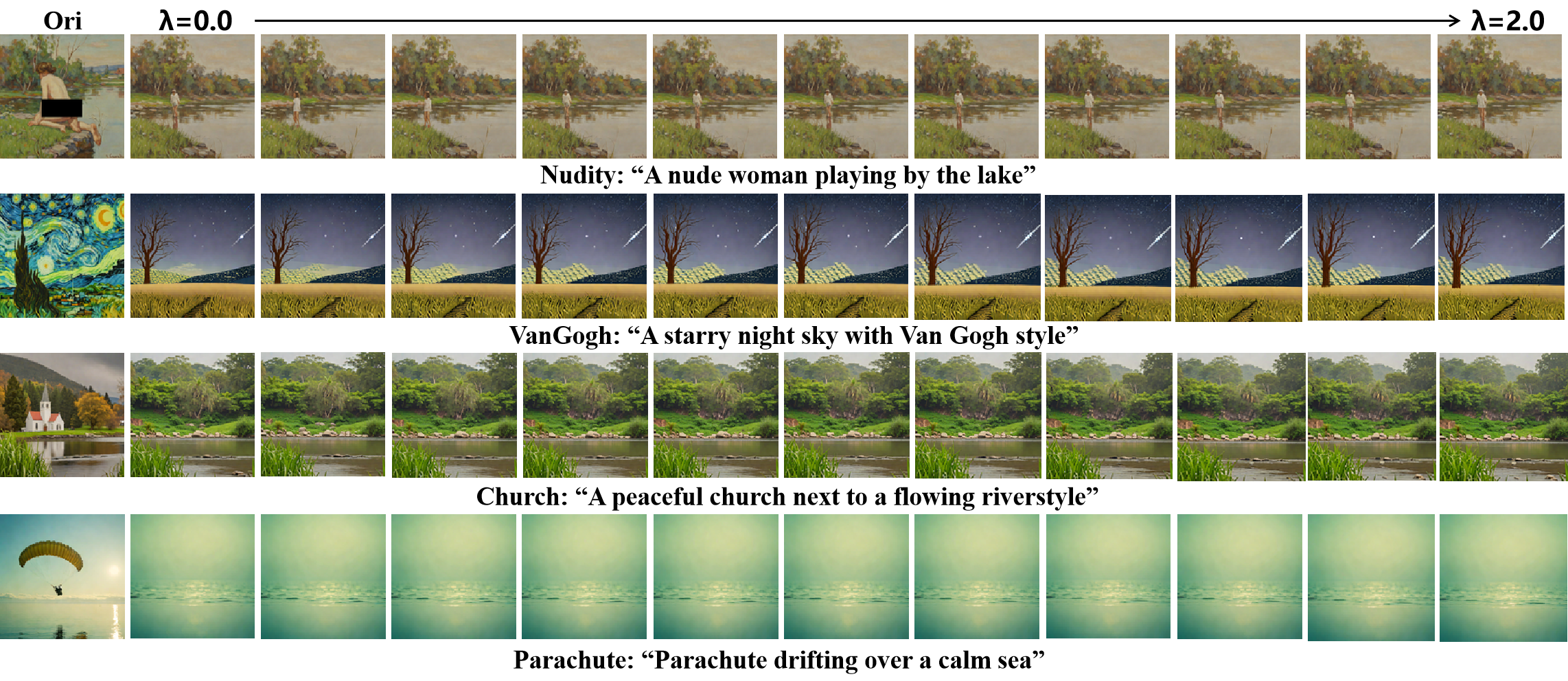}
    \caption{Forgetting effect across four unlearning tasks under different $\lambda$ values with a=-5. Generation step: 4.
}
    \label{fig:lambda_effect}
\end{figure*}
\begin{figure*}[t]
    \centering
    \renewcommand{\arraystretch}{1.2}
    \setlength{\tabcolsep}{2pt}
    \begin{tabular}
    {m{1.5cm}<{\centering}|ccccccccc}
        & Ori & $a=-3$ & $a=-4$ & $a=-5$ & $a=-6$ & $a=-7$ & $a=-8$ & $a=-9$ & $a=-10$ \\ \hline

        Nudity &
        \includegraphics[width=0.08\textwidth]{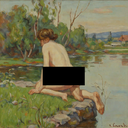} &
        \includegraphics[width=0.08\textwidth]{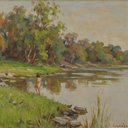} &
        \includegraphics[width=0.08\textwidth]{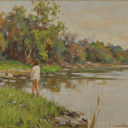} &
        \includegraphics[width=0.08\textwidth]{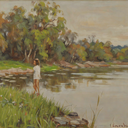} &
        \includegraphics[width=0.08\textwidth]{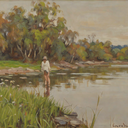} &
        \includegraphics[width=0.08\textwidth]{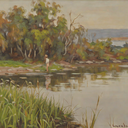} &
        \includegraphics[width=0.08\textwidth]{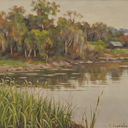} &
        \includegraphics[width=0.08\textwidth]{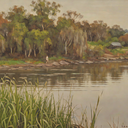} &
        \includegraphics[width=0.08\textwidth]{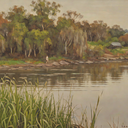} \\
        
        VanGogh &
        \includegraphics[width=0.08\textwidth]{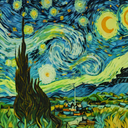} &
        \includegraphics[width=0.08\textwidth]{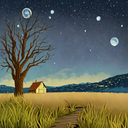} &
        \includegraphics[width=0.08\textwidth]{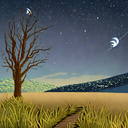} &
        \includegraphics[width=0.08\textwidth]{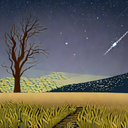} &
        \includegraphics[width=0.08\textwidth]{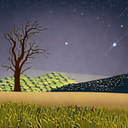} &
        \includegraphics[width=0.08\textwidth]{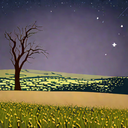} &
        \includegraphics[width=0.08\textwidth]{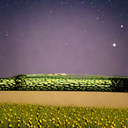} &
        \includegraphics[width=0.08\textwidth]{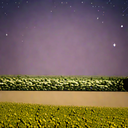} &
        \includegraphics[width=0.08\textwidth]{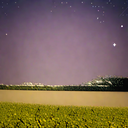} \\
        
        Church &
        \includegraphics[width=0.08\textwidth]{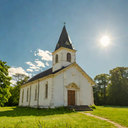} &
        \includegraphics[width=0.08\textwidth]{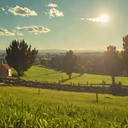} &
        \includegraphics[width=0.08\textwidth]{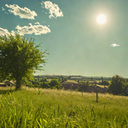} &
        \includegraphics[width=0.08\textwidth]{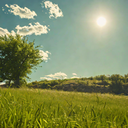} &
        \includegraphics[width=0.08\textwidth]{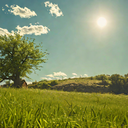} &
        \includegraphics[width=0.08\textwidth]{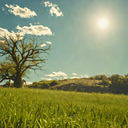} &
        \includegraphics[width=0.08\textwidth]{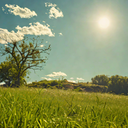} &
        \includegraphics[width=0.08\textwidth]{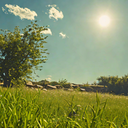} &
        \includegraphics[width=0.08\textwidth]{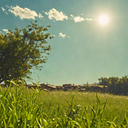} \\
        
        Parachute &
        \includegraphics[width=0.08\textwidth]{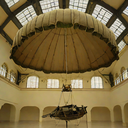} &
        \includegraphics[width=0.08\textwidth]{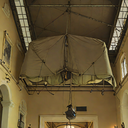} &
        \includegraphics[width=0.08\textwidth]{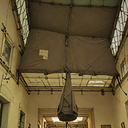} &
        \includegraphics[width=0.08\textwidth]{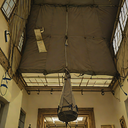} &
        \includegraphics[width=0.08\textwidth]{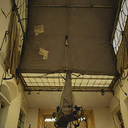} &
        \includegraphics[width=0.08\textwidth]{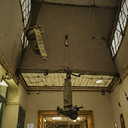} &
        \includegraphics[width=0.08\textwidth]{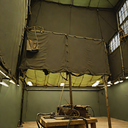} &
        \includegraphics[width=0.08\textwidth]{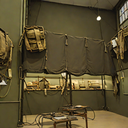} &
        \includegraphics[width=0.08\textwidth]{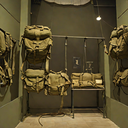} \\
    \end{tabular}
    
    \caption{Forgetting effect across four unlearning tasks under different $a$ values. 
    Rows correspond to different categories, while columns correspond to $a$ values. Generation steps: 4.}
    \label{fig:scale_effect}
\end{figure*}
\noindent\textbf{Effect of the Regularization Parameter $\lambda$.}
In Figure~\ref{fig:lambda_effect}, we examine the effect of the regularization coefficient \(\lambda\) on forgetting with the scaling factor fixed at \(a = -5\). Visual differences across different values of \(\lambda\) are subtle. Therefore, the optimal \(\lambda\) for the four forgetting tasks is selected based on quantitative metrics.

\noindent\textbf{Effect of the Scaling Factor $a$.} For the scaling factor \(a\), we vary it from \(-3\) to \(-10\) while fixing \(\lambda\) at its optimum, as shown in Figure~\ref{fig:scale_effect}. Small magnitudes, such as, \(a=-3\), yield weaker removal of the target concept; for instance, Van Gogh style retains characteristic stars patterns. Increasing $\left | a \right |$ strengthens suppression, and for nudity the person may disappear. However, larger magnitudes can degrade overall image quality. In our experiments, we observe that suppression factors approaching \(-10\) noticeably reduce aesthetic quality, which we hypothesize arises from imbalance due to overly strong feature activation or suppression. We therefore do not recommend using \(a< -10\).

\noindent\textbf{Efficiency and Resource Overhead.} For the efficiency and resource overhead, we evaluate our fine-tuning pipeline on a single NVIDIA A800 GPU (80GB PCIe) with CUDA 12.5 and NVIDIA driver version 555.42.06.
Constructing the fine-tuning dataset for 50 prompts takes only 14.07 seconds, which shows that preparing the dataset is highly efficient and does not introduce noticeable overhead.
Using precomputed latent input-output pairs, fine-tuning 50 samples for 100 epochs takes 670.3 seconds (11.17 minutes), averaging 6.7 seconds per epoch, excluding data loading and I/O overhead. For comparison, using prompt-based input requires 1,249.7 seconds (20.83 minutes) for the same dataset and epochs, averaging 12.5 seconds per epoch. Latent I/O fine-tuning bypasses the inference stage of the diffusion model and therefore avoids storing intermediate activations, resulting in significantly lower GPU memory consumption and enabling a larger batch size under the same hardware budget. We further observe that enabling automatic mixed-precision (AMP) training reduces the average per-epoch fine-tuning time from 6.7 seconds to 1.34 seconds on SDXL-Turbo and 0.63 seconds on SD v1.4, demonstrating additional efficiency gains. Although the latent-based dataset consumes more storage (71–72 MB) than the prompt-based dataset (28–29 MB), the reduction in fine-tuning time demonstrates a significant efficiency gain. In Table~\ref{tab:efficiency}, we compare our method with other baselines in terms of modification time, parameter changes and inference time. Notably, SAEUron, which performs unlearning via an external SAE module, incurs a 73.3\% increase in inference overhead on SDXL-Turbo. In contrast, although our fine-tuning-based approach requires slightly more time upfront to adapt the model, it introduces no additional inference cost. Overall, while maintaining strong forgetting performance, the time overhead of our method remains practical, requiring less than three minutes to forget a single concept.




\section{Relevant Features}
\label{appendix:relevantfeatures}
\begin{table}[t]
\centering
\small
\caption{The related features for the four main concepts identified by \textsc{\saedit} on SDXL-Turbo and SDXL.}
\label{tb:features}
\renewcommand{\arraystretch}{1.2}
\setlength{\tabcolsep}{4pt} 
\begin{tabular}{@{}p{1.5cm}|p{1.5cm}|p{1.5cm} |p{3cm}@{}} 
\toprule
\textbf{Model} & \textbf{Concept} & \textbf{Location} & \textbf{Feature IDs} \\
\midrule
\multirow{4}{*}{\shortstack{SDXL\\-Turbo}} 
& Nudity & down.2.1 & \textit{\#4913}, \textit{\#4001} \\
& VanGogh & up.0.1 & \textit{\#3166}, \textit{\#3950} \\
& Church & down.2.1 & \textit{\#3799}, \textit{\#2495}, \textit{\#2005}, \textit{\#2831}, \textit{\#3551}, \textit{\#4256}, \textit{\#4008}, \textit{\#1292} \\
& Parachute & down.2.1 & \textit{\#642}, \textit{\#3260}, \textit{\#4525}, \textit{\#3453}, \textit{\#4241}, \textit{\#3851}, \textit{\#4966}, \textit{\#4882}, \textit{\#4764} \\
\midrule
\multirow{4}{*}{SDXL} 
& Nudity & down.2.1 & \textit{\#4913}, \textit{\#4001} \\ 
& VanGogh & up.0.1 & \textit{\#4950}, \textit{\#3166}, \textit{\#2008}, \textit{\#4197}, \textit{\#1566}\\
& Church & down.2.1 & \textit{\#3799}, \textit{\#2005}, \textit{\#3551}, \textit{\#2495}, \textit{\#4008} , \textit{\#2831}, \textit{\#3921}\\
& Parachute & down.2.1 & \textit{\#642}, \textit{\#4525}, \textit{\#3260}, \textit{\#3453}, \textit{\#4882}, \textit{\#4764}, \textit{\#3851} , \textit{\#1712}, \textit{\#4966}\\
\bottomrule
\end{tabular}
\end{table}
In Table~\ref{tb:features}, we present the cross-attention blocks and feature indices associated with four representative concepts in both the SDXL-Turbo and SDXL models. The SAEs used for these two models are shared~\cite{surkov2025onestepenoughsparseautoencoders}, and the same dataset is employed when we identify the features corresponding to each concept. 

For \textit{Nudity}, \textit{Church} and \textit{Parachute}, the highly activated features in the same cross-attention blocks are largely consistent across the two models, while for \textit{VanGogh}, despite one overlapping feature, noticeable differences are observed.

In addition, \textit{Nudity} and \textit{VanGogh} involve relatively few related features, whereas \textit{Church} and \textit{Parachute} are associated with a larger number. 
We speculate that, unlike abstract concepts such as \textit{Nudity} or \textit{VanGogh}, concepts like \textit{Church} consist of multiple distinct visual components (e.g., spires, walls), which naturally lead to a larger set of associated features. The complexity and diversity of these visual elements make such objects harder to forget without inadvertently affecting unrelated concepts, which also explains why \textit{Church} and \textit{Parachute} usually serve as representative concepts for evaluating object-level forgetting. In addition, the lack of sufficiently diverse concept-relevant samples in the dataset used to train the SAEs may further prevent these features from being well concentrated. This observation provides an interesting direction for future exploration.

\begin{figure*}[ht]
    \centering
    \renewcommand{\arraystretch}{0.8}
    \setlength{\tabcolsep}{1.5pt}

    \begin{tabular}{c|cccccc|cccccc}
        & \multicolumn{6}{c|}{\textbf{T1}} & \multicolumn{6}{c}{\textbf{T4}} \\
&\shortstack{\small SDXL-\\Turbo} &
\raisebox{1ex}{\shortstack{\small ESD}} &
\raisebox{1ex}{\shortstack{\small UCE}}  & \small \shortstack{Concept\\Prune} & \small \shortstack{SAE\\Uron} & \shortstack{Concept\\ACP}
&\shortstack{\small SDXL-\\Turbo} &
\raisebox{1ex}{\shortstack{\small ESD}} &
\raisebox{1ex}{\shortstack{\small UCE}}  & \small \shortstack{Concept\\Prune} & \small \shortstack{SAE\\Uron} & \shortstack{Concept\\ACP}\\\hline

        \small \textbf{Nudity} &
        \includegraphics[width=0.07\textwidth]{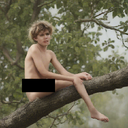} &
        \includegraphics[width=0.07\textwidth]{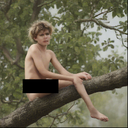} &
        \includegraphics[width=0.07\textwidth]{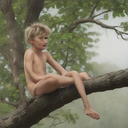} &
        \includegraphics[width=0.07\textwidth]{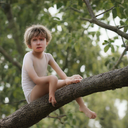} &
        \includegraphics[width=0.07\textwidth]{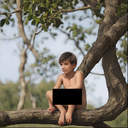} &
        \includegraphics[width=0.07\textwidth]{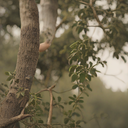} &
        \includegraphics[width=0.07\textwidth]{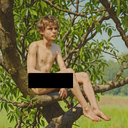} &
        \includegraphics[width=0.07\textwidth]{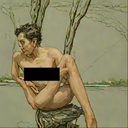} &
        \includegraphics[width=0.07\textwidth]{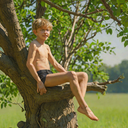} &
        \includegraphics[width=0.07\textwidth]{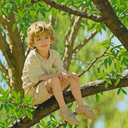} &
        \includegraphics[width=0.07\textwidth]{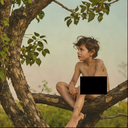} &
        \includegraphics[width=0.07\textwidth]{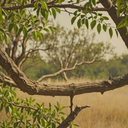} \\
        &
        \includegraphics[width=0.07\textwidth]{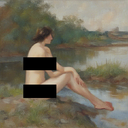} &
        \includegraphics[width=0.07\textwidth]{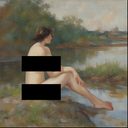} &
        \includegraphics[width=0.07\textwidth]{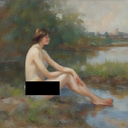} &
        \includegraphics[width=0.07\textwidth]{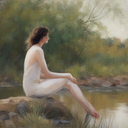} &
        \includegraphics[width=0.07\textwidth]{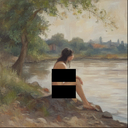} &
        \includegraphics[width=0.07\textwidth]{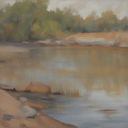} &
        \includegraphics[width=0.07\textwidth]{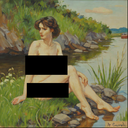} &
        \includegraphics[width=0.07\textwidth]{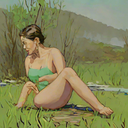} &
        \includegraphics[width=0.07\textwidth]{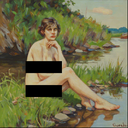} &
        \includegraphics[width=0.07\textwidth]{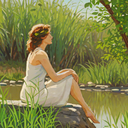} &
        \includegraphics[width=0.07\textwidth]{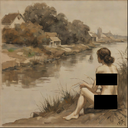} &
        \includegraphics[width=0.07\textwidth]{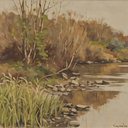} \\
        &
        \includegraphics[width=0.07\textwidth]{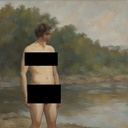} &
        \includegraphics[width=0.07\textwidth]{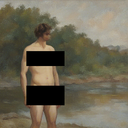} &
        \includegraphics[width=0.07\textwidth]{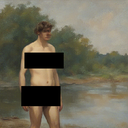} &
        \includegraphics[width=0.07\textwidth]{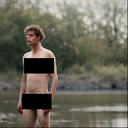} &
        \includegraphics[width=0.07\textwidth]{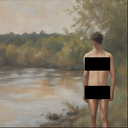} &
        \includegraphics[width=0.07\textwidth]{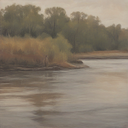} &
        \includegraphics[width=0.07\textwidth]{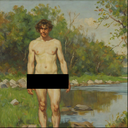} &
        \includegraphics[width=0.07\textwidth]{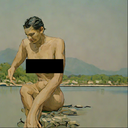} &
        \includegraphics[width=0.07\textwidth]{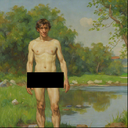} &
        \includegraphics[width=0.07\textwidth]{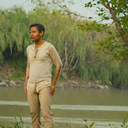} &
        \includegraphics[width=0.07\textwidth]{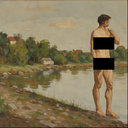} &
        \includegraphics[width=0.07\textwidth]{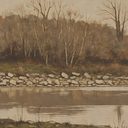} \\
        &
        \includegraphics[width=0.07\textwidth]{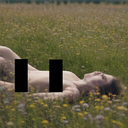} &
        \includegraphics[width=0.07\textwidth]{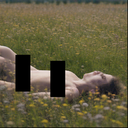} &
        \includegraphics[width=0.07\textwidth]{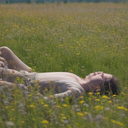} &
        \includegraphics[width=0.07\textwidth]{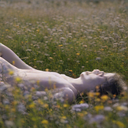} &
        \includegraphics[width=0.07\textwidth]{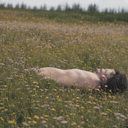} &
        \includegraphics[width=0.07\textwidth]{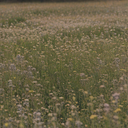} &
        \includegraphics[width=0.07\textwidth]{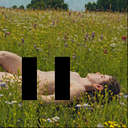} &
        \includegraphics[width=0.07\textwidth]{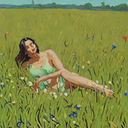} &
        \includegraphics[width=0.07\textwidth]{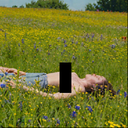} &
        \includegraphics[width=0.07\textwidth]{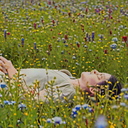} &
        \includegraphics[width=0.07\textwidth]{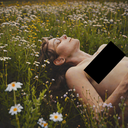} &
        \includegraphics[width=0.07\textwidth]{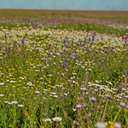} \\\hline

        \small\textbf{VanGogh} &
        \includegraphics[width=0.07\textwidth]{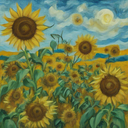} &
        \includegraphics[width=0.07\textwidth]{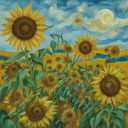} &
        \includegraphics[width=0.07\textwidth]{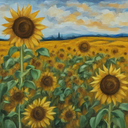} &
        \includegraphics[width=0.07\textwidth]{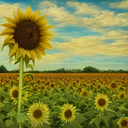} &
        \includegraphics[width=0.07\textwidth]{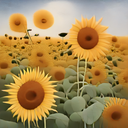} &
        \includegraphics[width=0.07\textwidth]{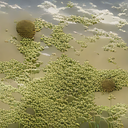} &
        \includegraphics[width=0.07\textwidth]{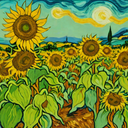} &
        \includegraphics[width=0.07\textwidth]{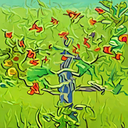} &
        \includegraphics[width=0.07\textwidth]{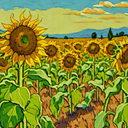} &
        \includegraphics[width=0.07\textwidth]{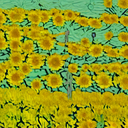} &
        \includegraphics[width=0.07\textwidth]{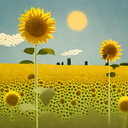} &
        \includegraphics[width=0.07\textwidth]{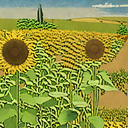} \\
        &
        \includegraphics[width=0.07\textwidth]{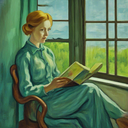} &
        \includegraphics[width=0.07\textwidth]{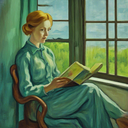} &
        \includegraphics[width=0.07\textwidth]{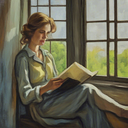} &
        \includegraphics[width=0.07\textwidth]{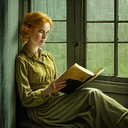} &
        \includegraphics[width=0.07\textwidth]{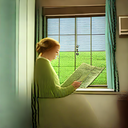} &
        \includegraphics[width=0.07\textwidth]{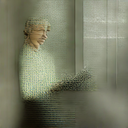} &
        \includegraphics[width=0.07\textwidth]{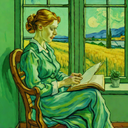} &
        \includegraphics[width=0.07\textwidth]{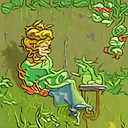} &
        \includegraphics[width=0.07\textwidth]{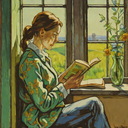} &
        \includegraphics[width=0.07\textwidth]{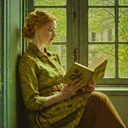} &
        \includegraphics[width=0.07\textwidth]{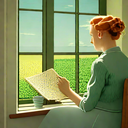} &
        \includegraphics[width=0.07\textwidth]{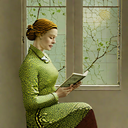} \\
        &
        \includegraphics[width=0.07\textwidth]{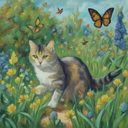} &
        \includegraphics[width=0.07\textwidth]{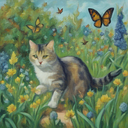} &
        \includegraphics[width=0.07\textwidth]{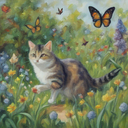} &
        \includegraphics[width=0.07\textwidth]{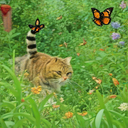} &
        \includegraphics[width=0.07\textwidth]{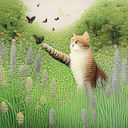} &
        \includegraphics[width=0.07\textwidth]{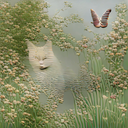} &
        \includegraphics[width=0.07\textwidth]{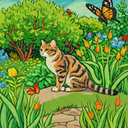} &
        \includegraphics[width=0.07\textwidth]{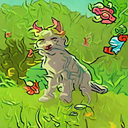} &
        \includegraphics[width=0.07\textwidth]{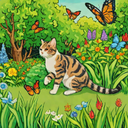} &
        \includegraphics[width=0.07\textwidth]{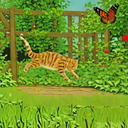} &
        \includegraphics[width=0.07\textwidth]{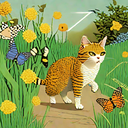} &
        \includegraphics[width=0.07\textwidth]{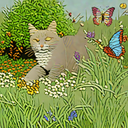} \\
        &
        \includegraphics[width=0.07\textwidth]{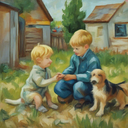} &
        \includegraphics[width=0.07\textwidth]{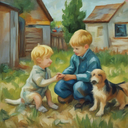} &
        \includegraphics[width=0.07\textwidth]{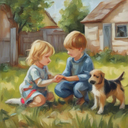} &
        \includegraphics[width=0.07\textwidth]{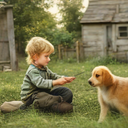} &
        \includegraphics[width=0.07\textwidth]{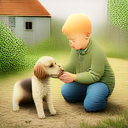} &
        \includegraphics[width=0.07\textwidth]{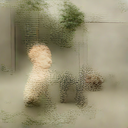} &
        \includegraphics[width=0.07\textwidth]{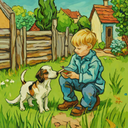} &
        \includegraphics[width=0.07\textwidth]{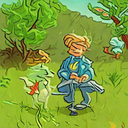} &
        \includegraphics[width=0.07\textwidth]{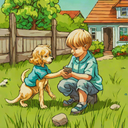} &
        \includegraphics[width=0.07\textwidth]{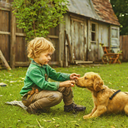} &
        \includegraphics[width=0.07\textwidth]{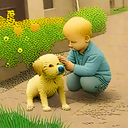} &
        \includegraphics[width=0.07\textwidth]{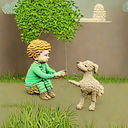} \\\hline

        \small\textbf{Church} &
        \includegraphics[width=0.07\textwidth]{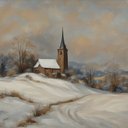} &
        \includegraphics[width=0.07\textwidth]{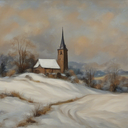} &
        \includegraphics[width=0.07\textwidth]{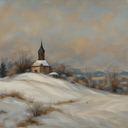} &
        \includegraphics[width=0.07\textwidth]{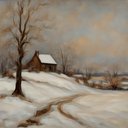} &
        \includegraphics[width=0.07\textwidth]{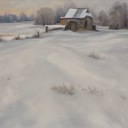} &
        \includegraphics[width=0.07\textwidth]{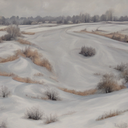} &
        \includegraphics[width=0.07\textwidth]{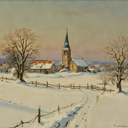} &
        \includegraphics[width=0.07\textwidth]{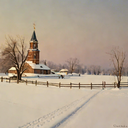} &
        \includegraphics[width=0.07\textwidth]{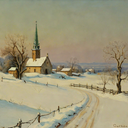} &
        \includegraphics[width=0.07\textwidth]{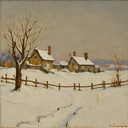} &
        \includegraphics[width=0.07\textwidth]{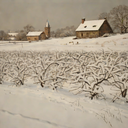} &
        \includegraphics[width=0.07\textwidth]{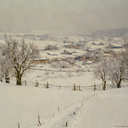} \\
        & 
        \includegraphics[width=0.07\textwidth]{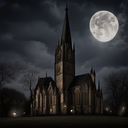} &
        \includegraphics[width=0.07\textwidth]{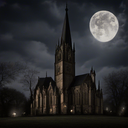} &
        \includegraphics[width=0.07\textwidth]{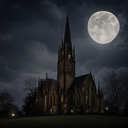} &
        \includegraphics[width=0.07\textwidth]{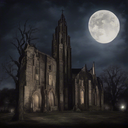} &
        \includegraphics[width=0.07\textwidth]{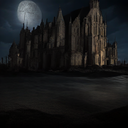} &
        \includegraphics[width=0.07\textwidth]{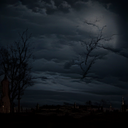}& 
        \includegraphics[width=0.07\textwidth]{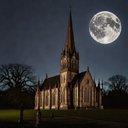} &
        \includegraphics[width=0.07\textwidth]{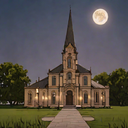} &
        \includegraphics[width=0.07\textwidth]{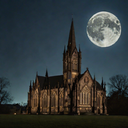} &
        \includegraphics[width=0.07\textwidth]{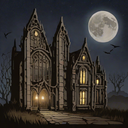} &
        \includegraphics[width=0.07\textwidth]{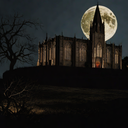} &
        \includegraphics[width=0.07\textwidth]{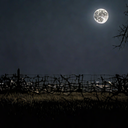} \\
        & 
        \includegraphics[width=0.07\textwidth]{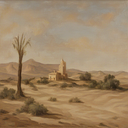} &
        \includegraphics[width=0.07\textwidth]{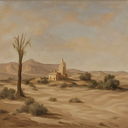} &
        \includegraphics[width=0.07\textwidth]{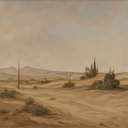} &
        \includegraphics[width=0.07\textwidth]{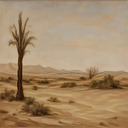} &
        \includegraphics[width=0.07\textwidth]{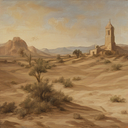} &
        \includegraphics[width=0.07\textwidth]{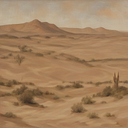}& 
        \includegraphics[width=0.07\textwidth]{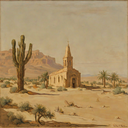} &
        \includegraphics[width=0.07\textwidth]{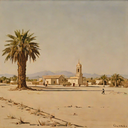} &
        \includegraphics[width=0.07\textwidth]{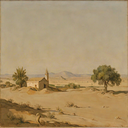} &
        \includegraphics[width=0.07\textwidth]{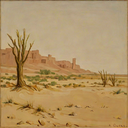} &
        \includegraphics[width=0.07\textwidth]{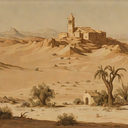} &
        \includegraphics[width=0.07\textwidth]{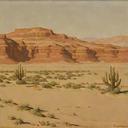} \\
        & 
        \includegraphics[width=0.07\textwidth]{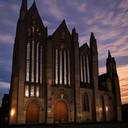} &
        \includegraphics[width=0.07\textwidth]{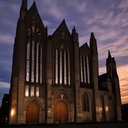} &
        \includegraphics[width=0.07\textwidth]{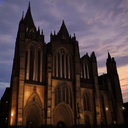} &
        \includegraphics[width=0.07\textwidth]{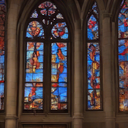} &
        \includegraphics[width=0.07\textwidth]{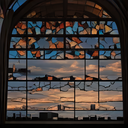} &
        \includegraphics[width=0.07\textwidth]{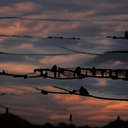} & 
        \includegraphics[width=0.07\textwidth]{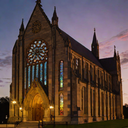} &
        \includegraphics[width=0.07\textwidth]{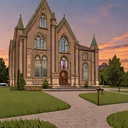} &
        \includegraphics[width=0.07\textwidth]{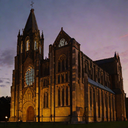} &
        \includegraphics[width=0.07\textwidth]{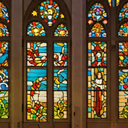} &
        \includegraphics[width=0.07\textwidth]{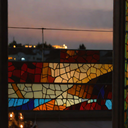} &
        \includegraphics[width=0.07\textwidth]{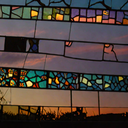}\\\hline

        \small\textbf{Parachute} &
        \includegraphics[width=0.07\textwidth]{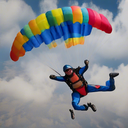} &
        \includegraphics[width=0.07\textwidth]{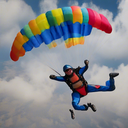} &
        \includegraphics[width=0.07\textwidth]{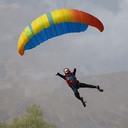} &
        \includegraphics[width=0.07\textwidth]{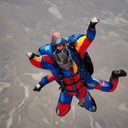} &
        \includegraphics[width=0.07\textwidth]{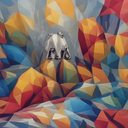} &
        \includegraphics[width=0.07\textwidth]{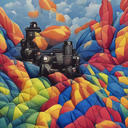} & 
        \includegraphics[width=0.07\textwidth]{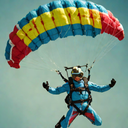} &
        \includegraphics[width=0.07\textwidth]{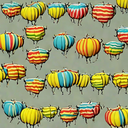} &
        \includegraphics[width=0.07\textwidth]{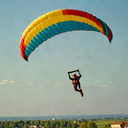} &
        \includegraphics[width=0.07\textwidth]{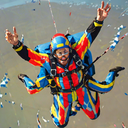} &
        \includegraphics[width=0.07\textwidth]{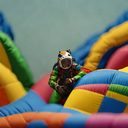} &
        \includegraphics[width=0.07\textwidth]{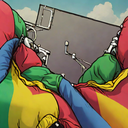} \\
        & 
        \includegraphics[width=0.07\textwidth]{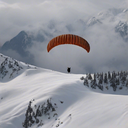} &
        \includegraphics[width=0.07\textwidth]{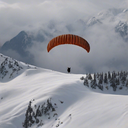} &
        \includegraphics[width=0.07\textwidth]{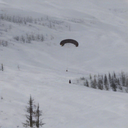} &
        \includegraphics[width=0.07\textwidth]{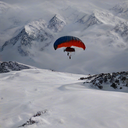} &
        \includegraphics[width=0.07\textwidth]{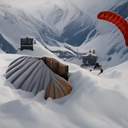} &
        \includegraphics[width=0.07\textwidth]{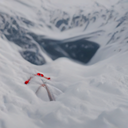}  & 
        \includegraphics[width=0.07\textwidth]{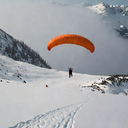} &
        \includegraphics[width=0.07\textwidth]{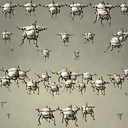} &
        \includegraphics[width=0.07\textwidth]{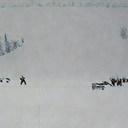} &
        \includegraphics[width=0.07\textwidth]{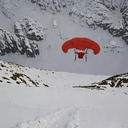} &
        \includegraphics[width=0.07\textwidth]{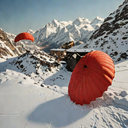} &
        \includegraphics[width=0.07\textwidth]{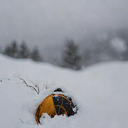}\\
        & 
        \includegraphics[width=0.07\textwidth]{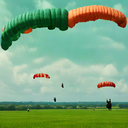} &
        \includegraphics[width=0.07\textwidth]{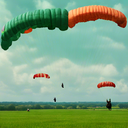} &
        \includegraphics[width=0.07\textwidth]{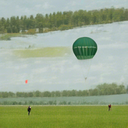} &
        \includegraphics[width=0.07\textwidth]{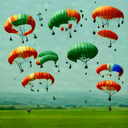} &
        \includegraphics[width=0.07\textwidth]{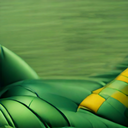} &
        \includegraphics[width=0.07\textwidth]{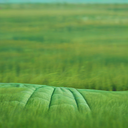}& 
        \includegraphics[width=0.07\textwidth]{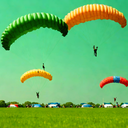} &
        \includegraphics[width=0.07\textwidth]{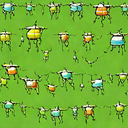} &
        \includegraphics[width=0.07\textwidth]{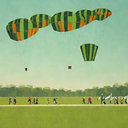} &
        \includegraphics[width=0.07\textwidth]{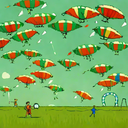} &
        \includegraphics[width=0.07\textwidth]{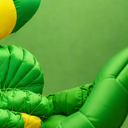} &
        \includegraphics[width=0.07\textwidth]{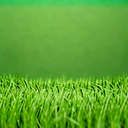} \\
        & 
        \includegraphics[width=0.07\textwidth]{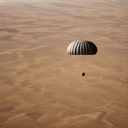} &
        \includegraphics[width=0.07\textwidth]{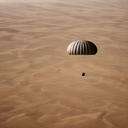} &
        \includegraphics[width=0.07\textwidth]{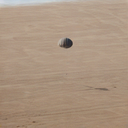} &
        \includegraphics[width=0.07\textwidth]{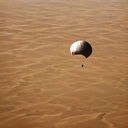} &
        \includegraphics[width=0.07\textwidth]{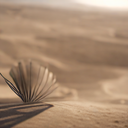} &
        \includegraphics[width=0.07\textwidth]{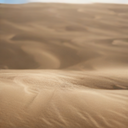} & 
        \includegraphics[width=0.07\textwidth]{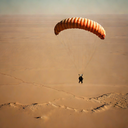} &
        \includegraphics[width=0.07\textwidth]{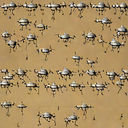} &
        \includegraphics[width=0.07\textwidth]{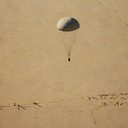} &
        \includegraphics[width=0.07\textwidth]{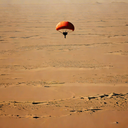} &
        \includegraphics[width=0.07\textwidth]{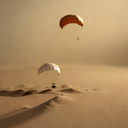} &
        \includegraphics[width=0.07\textwidth]{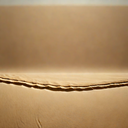}\\
        
    \end{tabular}

    \caption{Comparison of forgetting performance across four unlearning tasks under two generation steps.}
    \label{fig:baseline_all_t1_t4}
\end{figure*}

\end{document}